\documentclass[dvipsnames]{article} %
\usepackage{colm2024_conference}

\usepackage{booktabs}
\usepackage{graphicx}
\usepackage{enumitem}
\usepackage{wrapfig}
\usepackage{algorithm}
\usepackage{algpseudocode}

\usepackage{microtype}
\usepackage{amsmath}
\usepackage{colortbl}
\usepackage[utf8]{inputenc}
\definecolor{lightgray}{rgb}{0.9,0.9,0.9}
\usepackage{caption}
\usepackage{subcaption}
\usepackage{setspace}
\usepackage{url}
\usepackage{multirow}
\usepackage{colortbl}
\usepackage{tabularx}
\usepackage{blindtext}
\usepackage{pgfplots}
\pgfplotsset{compat=1.18} 
\usepackage{tikz}
\usetikzlibrary{er,positioning,bayesnet}
\usepackage{makecell}
\usepackage{tipa}
\usepackage{siunitx}
\usepackage{nicefrac}
\usepackage{tocloft}
\usepackage{listings}
\usepackage[raster,skins]{tcolorbox} %
\usepackage{xltabular}
\usepackage{adjustbox}
\usepackage{xurl}
\usepackage{rotating}
\usepackage[normalem]{ulem}
\useunder{\uline}{\ul}{}

\usepackage{amsmath,amsfonts,bm}

\def\eqref#1{equation~\ref{#1}}

\def\1{\bm{1}}

\DeclareMathAlphabet{\mathsfit}{\encodingdefault}{\sfdefault}{m}{sl}
\SetMathAlphabet{\mathsfit}{bold}{\encodingdefault}{\sfdefault}{bx}{n}

\renewcommand{\ghlink}{https://github.com/QwenLM/Qwen3}

\newcommand{\tabincell}[2]{\begin{tabular}{@{}#1@{}}#2\end{tabular}}

\newcommand*\justify{%
  \fontdimen2\font=0.4em
  \fontdimen3\font=0.2em
  \fontdimen4\font=0.1em
  \fontdimen7\font=0.1em
  \hyphenchar\font=`\-
}

\renewcommand{\texttt}[1]{%
  \begingroup
  \ttfamily
  \begingroup\lccode`~=`/\lowercase{\endgroup\def~}{/\discretionary{}{}{}}%
  \begingroup\lccode`~=`[\lowercase{\endgroup\def~}{[\discretionary{}{}{}}%
  \begingroup\lccode`~=`.\lowercase{\endgroup\def~}{.\discretionary{}{}{}}%
  \catcode`/=\active\catcode`[=\active\catcode`.=\active
  \justify\scantokens{#1\noexpand}%
  \endgroup
}

\title{Qwen3 Technical Report}

\author{
\bf Qwen Team
}

\begin{document}

\maketitle

\begin{abstract}
In this work, we present Qwen3, the latest version of the Qwen model family.
Qwen3 comprises a series of large language models (LLMs) designed to advance performance, efficiency, and multilingual capabilities. 
The Qwen3 series includes models of both dense and Mixture-of-Expert (MoE) architectures, with parameter scales ranging from 0.6 to 235 billion. 
A key innovation in Qwen3 is the integration of thinking mode (for complex, multi-step reasoning) and non-thinking mode (for rapid, context-driven responses) into a unified framework. 
This eliminates the need to switch between different models—--such as chat-optimized models (e.g., GPT-4o) and dedicated reasoning models (e.g., QwQ-32B)—--and enables dynamic mode switching based on user queries or chat templates.
Meanwhile, Qwen3 introduces a thinking budget mechanism, allowing users to allocate computational resources adaptively during inference, thereby balancing latency and performance based on task complexity.
Moreover, by leveraging the knowledge from the flagship models, we significantly reduce the computational resources required to build smaller-scale models, while ensuring their highly competitive performance.
Empirical evaluations demonstrate that Qwen3 achieves state-of-the-art results across diverse benchmarks, including tasks in code generation, mathematical reasoning, agent tasks, etc., competitive against larger MoE models and proprietary models. Compared to its predecessor Qwen2.5, Qwen3 expands multilingual support from 29 to 119 languages and dialects, enhancing global accessibility through improved cross-lingual understanding and generation capabilities.
To facilitate reproducibility and community-driven research and development, all Qwen3 models are publicly accessible under Apache 2.0. 

\end{abstract}

\vfill

\newpage

\section{Introduction}
\label{sec:intro}

The pursuit of artificial general intelligence (AGI) or artificial super intelligence (ASI) has long been a goal for humanity. Recent advancements in large foundation models, e.g., GPT-4o~\citep{gpt4o}, Claude 3.7~\citep{claude3.7}, Gemini 2.5~\citep{gemini2.5}, DeepSeek-V3~\citep{deepseekv3}, Llama-4~\citep{llama4}, and Qwen2.5~\citep{qwen2.5}, have demonstrated significant progress toward this objective. These models are trained on vast datasets spanning trillions of tokens across diverse domains and tasks, effectively distilling human knowledge and capabilities into their parameters. Furthermore, recent developments in reasoning models, optimized through reinforcement learning, highlight the potential for foundation models to enhance inference-time scaling and achieve higher levels of intelligence, e.g., o3~\citep{o3}, DeepSeek-R1~\citep{r1}. While most state-of-the-art models remain proprietary, the rapid growth of open-source communities has substantially reduced the performance gap between open-weight and closed-source models. Notably, an increasing number of top-tier models~\citep{llama4, deepseekv3, r1, qwen2.5} are now being released as open-source, fostering broader research and innovation in artificial intelligence.

In this work, we introduce Qwen3, the latest series in our foundation model family, Qwen. Qwen3 is a collection of open-weight large language models (LLMs) that achieve state-of-the-art performance across a wide variety of tasks and domains. We release both dense and Mixture-of-Experts (MoE) models, with the number of parameters ranging from 0.6 billion to 235 billion, to meet the needs of different downstream applications. Notably, the flagship model, Qwen3-235B-A22B, is an MoE model with a total of 235 billion parameters and 22 billion activated ones per token. This design ensures both high performance and efficient inference.

Qwen3 introduces several key advancements to enhance its functionality and usability. First, it integrates two distinct operating modes, thinking mode and non-thinking mode, into a single model. This allows users to switch between these modes without alternating between different models, e.g., switching from Qwen2.5 to QwQ~\citep{qwq}. This flexibility ensures that developers and users can adapt the model's behavior to suit specific tasks efficiently. Additionally, Qwen3 incorporates thinking budgets, providing users with fine-grained control over the level of reasoning effort applied by the model during task execution. This capability is crucial to the optimization of computational resources and performance, tailoring the model's thinking behavior to meet varying complexity in real-world applications. Furthermore, Qwen3 has been pre-trained on 36 trillion tokens covering up to 119 languages and dialects, effectively enhancing its multilingual capabilities. This broadened language support amplifies its potential for deployment in global use cases and international applications. 
These advancements together establish Qwen3 as a cutting-edge open-source large language model family, capable of effectively addressing complex tasks across various domains and languages.

The pre-training process for Qwen3 utilizes a large-scale dataset consisting of approximately 36 trillion tokens, curated to ensure linguistic and domain diversity. To efficiently expand the training data, we employ a multi-modal approach: Qwen2.5-VL~\citep{qwen2.5vl} is finetuned to extract text from extensive PDF documents. We also generate synthetic data using domain-specific models: Qwen2.5-Math~\citep{qwen2.5math} for mathematical content and Qwen2.5-Coder~\citep{qwen2.5coder} for code-related data. The pre-training process follows a three-stage strategy. In the first stage, the model is trained on about 30 trillion tokens to build a strong foundation of general knowledge. In the second stage, it is further trained on knowledge-intensive data to enhance reasoning abilities in areas like science, technology, engineering, and mathematics (STEM) and coding. Finally, in the third stage, the model is trained on long-context data to increase its maximum context length from 4,096 to 32,768 tokens.

To better align foundation models with human preferences and downstream applications, we employ a multi-stage post-training approach that empowers both thinking (reasoning) and non-thinking modes. In the first two stages, we focus on developing strong reasoning abilities through long chain-of-thought (CoT) cold-start finetuning and reinforcement learning focusing on mathematics and coding tasks. In the final two stages, we combine data with and without reasoning paths into a unified dataset for further fine-tuning, enabling the model to handle both types of input effectively, and we then apply general-domain reinforcement learning to improve performance across a wide range of downstream tasks. For smaller models, we use strong-to-weak distillation, leveraging both off-policy and on-policy knowledge transfer from larger models to enhance their capabilities.
Distillation from advanced teacher models significantly outperforms reinforcement learning in performance and training efficiency.

We evaluate both pre-trained and post-trained versions of our models across a comprehensive set of benchmarks spanning multiple tasks and domains. Experimental results show that our base pre-trained models achieve state-of-the-art performance. The post-trained models, whether in thinking or non-thinking mode, perform competitively against leading proprietary models and large mixture-of-experts (MoE) models such as o1, o3-mini, and DeepSeek-V3. Notably, our models excel in coding, mathematics, and agent-related tasks. For example, the flagship model Qwen3-235B-A22B achieves 85.7 on AIME'24 and 81.5 on AIME'25~\citep{aime}, 70.7 on LiveCodeBench v5~\citep{livecodebench}, 2,056 on CodeForces, and 70.8 on BFCL v3~\citep{bfcl}. In addition, other models in the Qwen3 series also show strong performance relative to their size. Furthermore, we observe that increasing the thinking budget for thinking tokens leads to a consistent improvement in the model's performance across various tasks.

In the following sections, we describe the design of the model architecture, provide details on its training procedures, present the experimental results of pre-trained and post-trained models, and finally, conclude this technical report by summarizing the key findings and outlining potential directions for future research.
\section{Architecture}

The Qwen3 series includes 6 dense models, namely Qwen3-0.6B, Qwen3-1.7B, Qwen3-4B, Qwen3-8B, Qwen3-14B, and Qwen3-32B, and 2 MoE models, Qwen3-30B-A3B and Qwen3-235B-A22B. The flagship model, Qwen3-235B-A22B, has a total of 235B parameters with 22B activated ones. Below, we elaborate on the architecture of the Qwen3 models.

The architecture of the Qwen3 dense models is similar to Qwen2.5~\citep{qwen2.5}, including using Grouped Query Attention (GQA, \citealp{gqa}), SwiGLU~\citep{glu}, Rotary Positional Embeddings (RoPE, \citealp{rope}), and RMSNorm~\citep{rmsnorm} with pre-normalization. Besides, we remove QKV-bias used in Qwen2~\citep{qwen2} and introduce QK-Norm~\citep{pmlr-v202-dehghani23a} to the attention mechanism to ensure stable training for Qwen3. Key information on model architecture is provided in Table~\ref{tab:arch-dense}.

The Qwen3 MoE models share the same fundamental architecture as the Qwen3 dense models. 
Key information on model architecture is provided in Table~\ref{tab:arch-moe}. 
We follow Qwen2.5-MoE~\citep{qwen2.5} and implement fine-grained expert segmentation~\citep{deepseekmoe}. 
The Qwen3 MoE models have 128 total experts with 8 activated experts per token.
Unlike Qwen2.5-MoE, the Qwen3-MoE design excludes shared experts.
Furthermore, we adopt the global-batch load balancing loss~\citep{global_balance} to encourage expert specialization.
These architectural and training innovations have yielded substantial improvements in model performance across downstream tasks.

Qwen3 models utilize Qwen's tokenizer~\citep{qwen}, which implements byte-level byte-pair encoding (BBPE,~\citealp{gpt3,wang2020neural,sennirch2016neural}) with a vocabulary size of 151,669.

\begin{table}[htbp]
\caption{Model architecture of Qwen3 dense models.\label{tab:arch-dense}}
\small
\centering
\begin{tabular}{@{}lcccc@{}} 
\toprule
Models  & Layers & Heads (Q / KV) & Tie Embedding & Context Length  \\
\midrule
Qwen3-0.6B  & 28 & 16 / 8 & Yes & 32K   \\
Qwen3-1.7B  & 28 & 16 / 8 & Yes & 32K   \\
Qwen3-4B  & 36 & 32 / 8 & Yes & 128K  \\
Qwen3-8B  & 36 & 32 / 8 & No & 128K  \\
Qwen3-14B & 40 & 40 / 8 & No & 128K  \\
Qwen3-32B & 64 & 64 / 8 & No & 128K  \\
\bottomrule
\end{tabular}
\end{table}

\begin{table}[htbp]
\caption{Model architecture of Qwen3 MoE models.\label{tab:arch-moe}}
\small
\centering
\begin{tabular}{@{}lcccc@{}} 
\toprule
Models  & Layers & Heads (Q / KV) & \# Experts (Total / Activated) & Context Length  \\
\midrule
Qwen3-30B-A3B  & 48 & 32 / 4 &128 / 8  & 128K   \\
Qwen3-235B-A22B  & 94 & 64 / 4 &128 / 8  & 128K  \\
\bottomrule
\end{tabular}
\end{table}

\section{Pre-training}

In this section, we describe the construction of our pretraining data, the details of our pretraining approach, and present experimental results from evaluating the base models on standard benchmarks.

\label{sec:pre}

\subsection{Pre-training Data}
Compared with Qwen2.5 \citep{qwen2.5}, we have significantly expanded the scale and diversity of our training data. Specifically, we collected twice as many pre-training tokens—covering three times more languages. All Qwen3 models are trained on a large and diverse dataset consisting of \textbf{119 languages and dialects}, with a total of \textbf{36 trillion tokens}. This dataset includes high-quality content in various domains such as coding, STEM (Science, Technology, Engineering, and Mathematics), reasoning tasks, books, multilingual texts, and synthetic data.

To further expand the pre-training data corpus, we first employ the Qwen2.5-VL model \citep{qwen2.5vl} to perform text recognition on a large volume of PDF-like documents. The recognized text is then refined using the Qwen2.5 model \citep{qwen2.5}, which helps improve its quality. Through this two-step process, we are able to obtain an additional set of high-quality text tokens, amounting to trillions in total. 
Besides, we employ Qwen2.5 \citep{qwen2.5}, Qwen2.5-Math \citep{qwen2.5math}, and Qwen2.5-Coder \citep{qwen2.5coder} models to synthesize trillions of text tokens in different formats, including textbooks, question-answering, instructions, and code snippets, covering dozens of domains. 
Finally, we further expand the pre-training corpus by incorporating additional multilingual data and introducing more languages. Compared to the pre-training data used in Qwen2.5, the number of supported languages has been significantly increased from 29 to 119, enhancing the model's linguistic coverage and cross-lingual capabilities.

We have developed a multilingual data annotation system designed to enhance both the quality and diversity of training data.
This system has been applied to our large-scale pre-training datasets, annotating over 30 trillion tokens across multiple dimensions such as educational value, fields, domains, and safety. These detailed annotations support more effective data filtering and combination. 
Unlike previous studies \citep{doremi,doge,regmix} that optimize the data mixture at the data source or domain level, our method optimizes the data mixture at the instance-level through extensive ablation experiments on small proxy models with the fine-grained data labels.

\subsection{Pre-training Stage}
The Qwen3 models are pre-trained through a three-stage process:

\begin{enumerate}[label=(\arabic*)]
    \item \textbf{General Stage (S1)}: At the first pre-training stage, all Qwen3 models are trained on over 30 trillion tokens using a sequence length of 4,096 tokens. At this stage, the models have been fully pre-trained on language proficiency and general world knowledge, with training data covering 119 languages and dialects.  
    \item \textbf{Reasoning Stage (S2)}: To further improve the reasoning ability, we optimize the pre-training corpus of this stage by increasing the proportion of STEM, coding, reasoning, and synthetic data. The models are further pre-trained with about 5T higher-quality tokens at a sequence length of 4,096 tokens. We also accelerate the learning rate decay during this stage.
    \item \textbf{Long Context Stage}: In the final pre-training stage, we collect high-quality long context corpora to extend the context length of Qwen3 models. All models are pre-trained on hundreds of billions of tokens with a sequence length of 32,768 tokens. The long context corpus includes 75\% of text between 16,384 to 32,768 tokens in length, and 25\% of text between 4,096 to 16,384 in length. Following Qwen2.5 \citep{qwen2.5}, we increase the base frequency of RoPE from 10,000 to 1,000,000 using the ABF technique \citep{ropeabf}. Meanwhile, we introduce YARN \citep{yarn} and Dual Chunk Attention~(DCA, \citealp{chunkllama}) to achieve a four-fold increase in sequence length capacity during inference.
\end{enumerate}

Similar to Qwen2.5 \citep{qwen2.5}, we develop scaling laws for optimal hyper-parameters (e.g., learning rate scheduler, and batch size) predictions based on three pre-training stages mentioned above. Through extensive experiments, we systematically study the relationship between model architecture, training data, training stage, and optimal training hyper-parameters. Finally, we set the predicted optimal learning rate and batch size strategy for each dense or MoE model.

\subsection{Pre-training Evaluation}

We conduct comprehensive evaluations of the base language models of the Qwen3 series. 
The evaluation of base models mainly focuses on their performance in general knowledge, reasoning, mathematics, scientific knowledge, coding, and multilingual capabilities. The evaluation datasets for pre-trained base models include 15 benchmarks:

\begin{itemize}

\item \textbf{General Tasks}: MMLU \citep{mmlu} (5-shot), MMLU-Pro \citep{mmlupro} (5-shot, CoT), MMLU-redux \citep{mmluredux} (5-shot), BBH \citep{bbh} (3-shot, CoT), SuperGPQA \citep{supergpqa}(5-shot, CoT).

\item \textbf{Math \& STEM Tasks}: GPQA \citep{gpqa} (5-shot, CoT), GSM8K \citep{gsm8k} (4-shot, CoT), MATH \citep{math} (4-shot, CoT).

\item \textbf{Coding Tasks}: EvalPlus \citep{evalplus} (0-shot) (Average of HumanEval \citep{humaneval}, MBPP \citep{mbpp}, Humaneval+, MBPP+) \citep{evalplus}, MultiPL-E \citep{multiple} (0-shot) (Python, C++, JAVA, PHP, TypeScript, C\#, Bash, JavaScript), MBPP-3shot \citep{mbpp}, CRUX-O of CRUXEval (1-shot) \citep{gu2024cruxeval}.

\item \textbf{Multilingual Tasks}: MGSM \citep{mgsm} (8-shot, CoT), MMMLU \citep{mmmlu} (5-shot), INCLUDE \citep{romanou2024includeevaluatingmultilinguallanguage} (5-shot).

\end{itemize}

For the base model baselines, we compare the Qwen3 series base models with the Qwen2.5 base models \citep{qwen2.5} and other leading open-source base models, including DeepSeek-V3 Base \citep{deepseekv3}, Gemma-3 \citep{gemma3}, Llama-3 \citep{llama3}, and Llama-4 \citep{llama4} series base models, in terms of scale of parameters. All models are evaluated using the same evaluation pipeline and the widely-used evaluation settings to ensure fair comparison.

\paragraph{Summary of Evaluation Results}
Based on the overall evaluation results, we highlight some key conclusions of Qwen3 base models. 
\begin{enumerate}[label=(\arabic*)]
\item Compared with the previously open-source SOTA dense and MoE base models (such as DeepSeek-V3 Base, Llama-4-Maverick Base, and Qwen2.5-72B-Base), Qwen3-235B-A22B-Base outperforms these models in most tasks with significantly fewer total parameters or activated parameters.

\item For the Qwen3 MoE base models, our experimental results indicate that: (a) Using the same pre-training data, Qwen3 MoE base models can achieve similar performance to Qwen3 dense base models with only \textbf{1/5} activated parameters. (b) Due to the improvements of the Qwen3 MoE architecture, the scale-up of the training tokens, and more advanced training strategies, the Qwen3 MoE base models can outperform the Qwen2.5 MoE base models with less than \textbf{1/2} activated parameters and fewer total parameters. (c) Even with \textbf{1/10} of the activated parameters of the Qwen2.5 dense base model, the Qwen3 MoE base model can achieve comparable performance, which brings us significant advantages in inference and training costs.

\item The overall performance of the Qwen3 dense base models is comparable to the Qwen2.5 base models at higher parameter scales. For example, Qwen3-1.7B/4B/8B/14B/32B-Base achieve comparable performance to Qwen2.5-3B/7B/14B/32B/72B-Base, respectively. Especially in STEM, coding, and reasoning benchmarks, the performance of Qwen3 dense base models even surpasses Qwen2.5 base models at higher parameter scales.
\end{enumerate}
The detailed results are as follows.

\begin{table}[tbp]
\centering
\caption{\textbf{Comparison among Qwen3-235B-A22B-Base and other representative strong open-source baselines. The highest,
the second-best scores are shown in \textbf{bold} and \underline{underlined}, respectively.}}
\label{tab:base-235A22B}
\small %
\setlength{\tabcolsep}{3pt} %
\begin{tabular}{@{}lccccc@{}}
\toprule
& \textbf{Qwen2.5-72B} & \textbf{Qwen2.5-Plus} & \textbf{Llama-4-Maverick} & \textbf{DeepSeek-V3} & \textbf{Qwen3-235B-A22B} \\
& \textbf{Base} & \textbf{Base} & \textbf{Base} & \textbf{Base} & \textbf{Base} \\
\midrule
Architecture & Dense & MoE & MoE & MoE & MoE\\
\# Total Params & 72B & 271B & 402B & 671B & 235B\\
\# Activated Params & 72B & 37B & 17B & 37B & 22B\\
\midrule
\multicolumn{6}{c}{\textit{General Tasks}} \\
\midrule
MMLU & 86.06 & 85.02 & 85.16 & \underline{87.19} & \textbf{87.81}\\
MMLU-Redux & 83.91 & 82.69 & 84.05 & \underline{86.14} & \textbf{87.40} \\
MMLU-Pro & 58.07 & 63.52 & \underline{63.91} & 59.84 & \textbf{68.18} \\
SuperGPQA & 36.20 & 37.18 & 40.85 & \underline{41.53} & \textbf{44.06} \\
BBH & \underline{86.30} & 85.60 & 83.62 & 86.22 & \textbf{88.87} \\
\midrule
\multicolumn{6}{c}{\textit{Math \& STEM Tasks}} \\
\midrule
GPQA & \underline{45.88} & 41.92 & 43.94 & 41.92 & \textbf{47.47} \\
GSM8K & 91.50 & \underline{91.89} & 87.72 & 87.57 & \textbf{94.39} \\
MATH & 62.12 & 62.78 & \underline{63.32} & 62.62 & \textbf{71.84} \\
\midrule
\multicolumn{6}{c}{\textit{Coding Tasks}} \\
\midrule
EvalPlus & 65.93 & 61.43 & \underline{68.38} & 63.75 & \textbf{77.60} \\
MultiPL-E & 58.70 & 62.16 & 57.28 & \underline{62.26} & \textbf{65.94} \\
MBPP & \underline{76.00} & 74.60 & 75.40 & 74.20 & \textbf{81.40} \\
CRUX-O & 66.20 & 68.50 & \underline{77.00} & 76.60 & \textbf{79.00} \\
\midrule
\multicolumn{6}{c}{\textit{Multilingual Tasks}} \\
\midrule
MGSM & 82.40 & 82.21 & 79.69 & \underline{82.68} & \textbf{83.53} \\
MMMLU & 84.40 & 83.49 & 83.09 & \underline{85.88} & \textbf{86.70} \\
INCLUDE & 69.05 & 66.97 & \underline{73.47} & \textbf{75.17} & 73.46 \\
\bottomrule
\end{tabular}
\end{table}

\paragraph{Qwen3-235B-A22B-Base}
We compare Qwen3-235B-A22B-Base to our previous similar-sized MoE Qwen2.5-Plus-Base \citep{qwen2.5} and other leading open-source base models: Llama-4-Maverick \citep{llama4}, Qwen2.5-72B-Base \citep{qwen2.5}, DeepSeek-V3 Base \citep{deepseekv3}. From the results in Table~\ref{tab:base-235A22B}, the Qwen3-235B-A22B-Base model attains the highest performance scores across most of the evaluated benchmarks. We further compare Qwen3-235B-A22B-Base with other baselines separately for the detailed analysis.
\begin{enumerate}[label=(\arabic*)]
    \item  Compared with the recently open-source model Llama-4-Maverick-Base, which has about \textbf{twice} the number of parameters, Qwen3-235B-A22B-Base still performs better on most benchmarks.
    \item  Compared with the previously state-of-the-art open-source model DeepSeek-V3-Base, Qwen3-235B-A22B-Base outperforms DeepSeek-V3-Base on 14 out of 15 evaluation benchmarks with only about \textbf{1/3} the total number of parameters and \textbf{2/3} activated parameters, demonstrating the powerful and cost-effectiveness of our models. 
    \item  Compared with our previous MoE Qwen2.5-Plus of similar size, Qwen3-235B-A22B-Base significantly outperforms it with fewer parameters and activated parameters, which shows the remarkable advantages of Qwen3 in pre-training data, training strategy, and model architecture.
    \item  Compared with our previous flagship open-source dense model Qwen2.5-72B-Base, Qwen3-235B-A22B-Base surpasses the latter in all benchmarks and uses fewer than \textbf{1/3} of the activated parameters. Meanwhile, due to the advantage of the model architecture, the inference costs and training costs on each trillion tokens of Qwen3-235B-A22B-Base are much cheaper than those of Qwen2.5-72B-Base.
\end{enumerate}

\begin{table}[tbp]
\centering
\caption{\textbf{Comparison among Qwen3-32B-Base and other strong open-source baselines. The highest and second-best scores are shown in \textbf{bold} and \underline{underlined}, respectively.}}
\label{tab:base-32B}
\small
\begin{tabular}{@{}lccccc@{}}
\toprule
 & \textbf{Qwen2.5-32B} & \textbf{Qwen2.5-72B} & \textbf{Gemma-3-27B} & \textbf{Llama-4-Scout} & \textbf{Qwen3-32B} \\
 & \textbf{Base} & \textbf{Base} & \textbf{Base} & \textbf{Base}& \textbf{Base} \\
\midrule
Architecture & Dense & Dense & Dense & MoE & Dense\\
\# Total Params & 32B & 72B & 27B & 109B & 32B\\
\# Activated Params & 32B & 72B & 27B & 17B & 32B\\
\midrule
\multicolumn{6}{c}{\textit{General Tasks}} \\
\midrule
MMLU & 83.32 & \textbf{86.06} & 78.69 & 78.27 & \underline{83.61} \\
MMLU-Redux & 81.97 & \textbf{83.91} & 76.53 & 71.09 & \underline{83.41} \\
MMLU-Pro & 55.10 & \underline{58.07} & 52.88 & 56.13 & \textbf{65.54} \\
SuperGPQA & 33.55 & \underline{36.20} & 29.87 & 26.51 & \textbf{39.78} \\
BBH & 84.48 & \underline{86.30} & 79.95 & 82.40 & \textbf{87.38} \\
\midrule
\multicolumn{6}{c}{\textit{Math \& STEM Tasks}} \\
\midrule
GPQA & \underline{47.97} & 45.88 & 26.26 & 40.40 & \textbf{49.49} \\
GSM8K & \underline{92.87} & 91.50 & 81.20 & 85.37 & \textbf{93.40} \\
MATH & 57.70 & \textbf{62.12} & 51.78 & 51.66 & \underline{61.62} \\
\midrule
\multicolumn{6}{c}{\textit{Coding Tasks}} \\
\midrule
EvalPlus & \underline{66.25} & 65.93 & 55.78 & 59.90 & \textbf{72.05} \\
MultiPL-E & 58.30 & \underline{58.70} & 45.03 & 47.38 & \textbf{67.06} \\
MBPP & 73.60 & \underline{76.00} & 68.40 & 68.60 & \textbf{78.20} \\
CRUX-O & \underline{67.80} & 66.20 & 60.00 & 61.90 & \textbf{72.50} \\
\midrule
\multicolumn{6}{c}{\textit{Multilingual Tasks}} \\
\midrule
MGSM & 78.12 & \underline{82.40} & 73.74 & 79.93 & \textbf{83.06} \\
MMMLU & 82.40 & \textbf{84.40} & 77.62 & 74.83 & \underline{83.83} \\
INCLUDE & 64.35 & \textbf{69.05} & \underline{68.94} & 68.09 & 67.87 \\
\bottomrule
\end{tabular}
\end{table}

\begin{table}[tbp]
\centering
\caption{\textbf{Comparison among Qwen3-14B-Base, Qwen3-30B-A3B-Base, and other strong open-source baselines. The highest and second-best scores are shown in \textbf{bold} and \underline{underlined}, respectively.}}
\label{tab:base-14B}
\adjustbox{center=\textwidth}{
\small
\setlength{\tabcolsep}{3pt} %
\begin{tabular}{@{}lcccccc@{}}
\toprule
 & \textbf{Gemma-3-12B} & \textbf{Qwen2.5-14B} & \textbf{Qwen2.5-32B} & \textbf{Qwen2.5-Turbo} & \textbf{Qwen3-14B} & \textbf{Qwen3-30B-A3B} \\
 &   \textbf{Base} & \textbf{Base} & \textbf{Base} & \textbf{Base}& \textbf{Base} & \textbf{Base}\\
\midrule
Architecture & Dense & Dense & Dense & MoE & Dense & MoE\\
\# Total Params & 12B & 14B & 32B & 42B & 14B & 30B\\
\# Activated Params & 12B & 14B & 32B & 6B & 14B & 3B\\
\midrule
\multicolumn{7}{c}{\textit{General Tasks}} \\
\midrule
MMLU & 73.87 & 79.66 & \textbf{83.32} & 79.50 & 81.05 & \underline{81.38} \\
MMLU-Redux & 70.70 & 76.64 & \textbf{81.97} & 77.11 & 79.88 & \underline{81.17} \\
MMLU-Pro & 44.91 & 51.16 & 55.10 & 55.60 & \underline{61.03} & \textbf{61.49} \\
SuperGPQA & 24.61 & 30.68 & 33.55 & 31.19 & \underline{34.27} & \textbf{35.72} \\
BBH & 74.28 & 78.18 & \textbf{84.48} & 76.10 & 81.07 & \underline{81.54} \\
\midrule
\multicolumn{7}{c}{\textit{Math \& STEM Tasks}} \\
\midrule
GPQA & 31.31 & 32.83 & \textbf{47.97} & 41.41 & 39.90 & \underline{43.94} \\
GSM8K & 78.01 & 90.22 & \textbf{92.87} & 88.32 & \underline{92.49} & 91.81 \\
MATH & 44.43 & 55.64 & 57.70 & 55.60 & \textbf{62.02} & \underline{59.04} \\
\midrule
\multicolumn{7}{c}{\textit{Coding Tasks}} \\
\midrule
EvalPlus & 52.65 & 60.70 & 66.25 & 61.23 & \textbf{72.23} & \underline{71.45} \\
MultiPL-E & 43.03 & 54.79 & 58.30 & 53.24 & \underline{61.69} & \textbf{66.53} \\
MBPP & 60.60 & 69.00 & \underline{73.60} & 67.60 & 73.40 & \textbf{74.40} \\
CRUX-O & 52.00 & 61.10 & \underline{67.80} & 60.20 & \textbf{68.60} & 67.20 \\
\midrule
\multicolumn{7}{c}{\textit{Multilingual Tasks}} \\
\midrule
MGSM & 64.35 & 74.68 & 78.12 & 70.45 & \textbf{79.20} & \underline{79.11} \\
MMMLU & 72.50 & 78.34 & \textbf{82.40} & 79.76 & 79.69 & \underline{81.46} \\
INCLUDE & 63.34 & 60.26 & 64.35 & 59.25 & \underline{64.55} & \textbf{67.00} \\
\bottomrule
\end{tabular}
}
\end{table}

\begin{table}[tbp]
\centering
\caption{\textbf{Comparison among Qwen8B-Base and other strong open-source baselines. The highest and second-best scores are shown in \textbf{bold} and \underline{underlined}, respectively.}}
\label{tab:base-8B}
\small
\begin{tabular}{@{}lcccc@{}}
\toprule
 & \textbf{Llama-3-8B} & \textbf{Qwen2.5-7B} & \textbf{Qwen2.5-14B}  & \textbf{Qwen3-8B} \\
&   \textbf{Base} & \textbf{Base} & \textbf{Base} & \textbf{Base} \\
\midrule
Architecture & Dense & Dense & Dense & Dense\\
\# Total Params & 8B & 7B & 14B & 8B \\
\# Activated Params & 8B & 7B & 14B & 8B \\
\midrule
\multicolumn{5}{c}{\textit{General Tasks}} \\
\midrule
MMLU & 66.60 & 74.16 & \textbf{79.66} & \underline{76.89}  \\
MMLU-Redux & 61.59 & 71.06 & \textbf{76.64} & \underline{76.17}  \\
MMLU-Pro & 35.36 & 45.00 &  \underline{51.16} & \textbf{56.73}  \\
SuperGPQA & 20.54 & 26.34 & \underline{30.68} & \textbf{31.64}  \\
BBH & 57.70 & 70.40 & \underline{78.18} & \textbf{78.40} \\
\midrule
\multicolumn{5}{c}{\textit{Math \& STEM Tasks}} \\
\midrule
GPQA & 25.80 & \underline{36.36} & 32.83 & \textbf{44.44} \\
GSM8K & 55.30 & 85.36 & \textbf{90.22} & \underline{89.84} \\
MATH & 20.50 & 49.80 & \underline{55.64} & \textbf{60.80}  \\
\midrule
\multicolumn{5}{c}{\textit{Coding Tasks}} \\
\midrule
EvalPlus & 44.13 & \underline{62.18} & 60.70 & \textbf{67.65} \\
MultiPL-E & 31.45 & 50.73 & \underline{54.79} & \textbf{58.75}  \\
MBPP & 48.40 & 63.40 & \underline{69.00} & \textbf{69.80} \\
CRUX-O & 36.80 & 48.50 & \underline{61.10} & \textbf{62.00}  \\
\midrule
\multicolumn{5}{c}{\textit{Multilingual Tasks}} \\
\midrule
MGSM & 38.92 & 63.60 & \underline{74.68} & \textbf{76.02}  \\
MMMLU & 59.65 & 71.34 & \textbf{78.34} & \underline{75.72} \\
IINCLUDE & 44.94 & 53.98 & \textbf{60.26} & \underline{59.40} \\
\bottomrule
\end{tabular}
\end{table}

\begin{table}[tbp]
\centering
\caption{\textbf{Comparison among Qwen3-4B-Base and other strong open-source baselines. The highest and second-best scores are shown in \textbf{bold} and \underline{underlined}, respectively.}}
\label{tab:base-4B}
\small %
\begin{tabular}{@{}lcccc@{}}
\toprule
 & \textbf{Gemma-3-4B} & \textbf{Qwen2.5-3B} & \textbf{Qwen2.5-7B} & \textbf{Qwen3-4B} \\
 & \textbf{Base} & \textbf{Base}& \textbf{Base}& \textbf{Base} \\
\midrule
Architecture & Dense & Dense & Dense & Dense\\
\# Total Params & 4B & 3B & 7B & 4B \\
\# Activated Params & 4B & 3B & 7B & 4B \\
\midrule
\multicolumn{5}{c}{\textit{General Tasks}} \\
\midrule
MMLU & 59.51 & 65.62 & \textbf{74.16} & \underline{72.99} \\
MMLU-Redux & 56.91 & 63.68 & \underline{71.06} & \textbf{72.79} \\
MMLU-Pro & 29.23 & 34.61 & \underline{45.00} & \textbf{50.58} \\
SuperGPQA & 17.68 & 20.31 & \underline{26.34} & \textbf{28.43} \\
BBH & 51.70 & 56.30 & \underline{70.40} & \textbf{72.59} \\
\midrule
\multicolumn{5}{c}{\textit{Math \& STEM Tasks}} \\
\midrule
GPQA & 24.24 & 26.26 & \underline{36.36} & \textbf{36.87} \\
GSM8K & 43.97 & 79.08 & \underline{85.36} & \textbf{87.79} \\
MATH & 26.10 & 42.64 & \underline{49.80} & \textbf{54.10} \\
\midrule
\multicolumn{5}{c}{\textit{Coding Tasks}} \\
\midrule
EvalPlus & 43.23 & 46.28 & \underline{62.18} & \textbf{63.53} \\
MultiPL-E & 28.06 & 39.65 & \underline{50.73} & \textbf{53.13} \\
MBPP & 46.40 & 54.60 & \underline{63.40} & \textbf{67.00} \\
CRUX-O & 34.00 & 36.50 & \underline{48.50} & \textbf{55.00} \\
\midrule
\multicolumn{5}{c}{\textit{Multilingual Tasks}} \\
\midrule
MGSM & 33.11 & 47.53 & \underline{63.60} & \textbf{67.74} \\
MMMLU & 59.62 & 65.55 & \underline{71.34} & \textbf{71.42} \\
INCLUDE & 49.06 & 45.90 & \underline{53.98} & \textbf{56.29} \\
\bottomrule
\end{tabular}
\end{table}

\begin{table}[tbp]
\centering
\caption{\textbf{Comparison among Qwen3-1.7B-Base, Qwen3-0.6B-Base, and other strong open-source baselines. The highest and second-best scores are shown in \textbf{bold} and \underline{underlined}, respectively.}}
\label{tab:base-1B}
\small %
\begin{tabular}{@{}lccccc@{}}
\toprule
 & \textbf{Qwen2.5-0.5B} & \textbf{Qwen3-0.6B} & \textbf{Gemma-3-1B} & \textbf{Qwen2.5-1.5B} & \textbf{Qwen3-1.7B} \\
 & \textbf{Base} & \textbf{Base} & \textbf{Base} & \textbf{Base} & \textbf{Base} \\
\midrule
Architecture & Dense & Dense & Dense & Dense & Dense\\
\# Total Params & 0.5B & 0.6B & 1B & 1.5B & 1.7B \\
\# Activated Params & 0.5B & 0.6B & 1B & 1.5B & 1.7B \\
\midrule
\multicolumn{6}{c}{\textit{General Tasks}} \\
\midrule
MMLU & 47.50 & 52.81 & 26.26 & \underline{60.90} & \textbf{62.63} \\
MMLU-Redux & 45.10 & 51.26 & 25.99 & \underline{58.46} & \textbf{61.66} \\
MMLU-Pro & 15.69 & 24.74 & 9.72 & \underline{28.53} & \textbf{36.76} \\
SuperGPQA & 11.30 & 15.03 & 7.19 & \underline{17.64} & \textbf{20.92} \\
BBH & 20.30 & 41.47 & 28.13 & \underline{45.10} & \textbf{54.47} \\
\midrule
\multicolumn{6}{c}{\textit{Math \& STEM Tasks}} \\
\midrule
GPQA & 24.75 & \underline{26.77} & 24.75 & 24.24 & \textbf{28.28} \\
GSM8K & 41.62 & 59.59 & 2.20 & \underline{68.54} & \textbf{75.44} \\
MATH & 19.48 & 32.44 & 3.66 & \underline{35.00} & \textbf{43.50} \\
\midrule
\multicolumn{6}{c}{\textit{Coding Tasks}} \\
\midrule
EvalPlus & 31.85 & 36.23 & 8.98 & \underline{44.80} & \textbf{52.70} \\
MultiPL-E & 18.70 & 24.58 & 5.15 & \underline{33.10} & \textbf{42.71} \\
MBPP & 29.80 & 36.60 & 9.20 & \underline{43.60} & \textbf{55.40} \\
CRUX-O & 12.10 & 27.00 & 3.80 & \underline{29.60} & \textbf{36.40} \\
\midrule
\multicolumn{6}{c}{\textit{Multilingual Tasks}} \\
\midrule
MGSM & 12.07 & 30.99 & 1.74 & \underline{32.82} & \textbf{50.71} \\
MMMLU & 31.53 & 50.16 & 26.57 & \underline{60.27} & \textbf{63.27} \\
INCLUDE & 24.74 & 34.26 & 25.62 & \underline{39.55} & \textbf{45.57} \\
\bottomrule
\end{tabular}
\end{table}

\paragraph{Qwen3-32B-Base}
Qwen3-32B-Base is our largest dense model among the Qwen3 series. We compare it to the baselines of similar sizes, including Gemma-3-27B \citep{gemma3} and Qwen2.5-32B \citep{qwen2.5}. In addition, we introduce two strong baselines: the recently open-source MoE model Llama-4-Scout, which has three times the parameters of Qwen3-32B-Base but half the activated parameters; and our previous flagship open-source dense model Qwen2.5-72B-Base, which has more than twice the number of parameters compared to Qwen3-32B-Base.
The results are shown in Table~\ref{tab:base-32B}, which support three key conclusions:
\begin{enumerate}[label=(\arabic*)]
\item Compared with the similar-sized models, Qwen3-32B-Base outperforms Qwen2.5-32B-Base and Gemma-3-27B Base on most benchmarks. Notably, Qwen3-32B-Base achieves 65.54 on MMLU-Pro and 39.78 on SuperGPQA, significantly outperforming its predecessor Qwen2.5-32B-Base. In addition, Qwen3-32B-Base achieves significantly higher encoding benchmark scores than all baseline models.
\item Surprisingly, we find that Qwen3-32B-Base achieves competitive results compared to Qwen2.5-72B-Base. Although Qwen3-32B-Base has less than half the number of parameters of Qwen2.5-72B-Base, it outperforms Qwen2.5-72B-Base in 10 of the 15 evaluation benchmarks. On coding, mathematics, and reasoning benchmarks, Qwen3-32B-Base has remarkable advantages.
\item Compared to Llama-4-Scout-Base, Qwen3-32B-Base significantly outperforms it on all 15 benchmarks, with only one-third of the number of parameters of Llama-4-Scout-Base, but twice the number of activated parameters.
\end{enumerate}

\paragraph{Qwen3-14B-Base \& Qwen3-30B-A3B-Base}
The evaluation of the Qwen3-14B-Base and Qwen3-30B-A3B-Base is compared against baselines of similar sizes, including Gemma-3-12B Base, Qwen2.5-14B Base. Similarly, we also introduce two strong baselines: (1) Qwen2.5-Turbo \citep{qwen2.5}, which has 42B parameters and 6B activated parameters. Note that its activated parameters are twice those of Qwen3-30B-A3B-Base. (2) Qwen2.5-32B-Base, which has 11 times the activated parameters of Qwen3-30B-A3B and more than twice that of Qwen3-14B.
The results are shown in Table~\ref{tab:base-14B}, where we can draw the following conclusions.
\begin{enumerate}[label=(\arabic*)]
\item Compared with the similar-sized models, Qwen3-14B-Base significantly performs better than Qwen2.5-14B-Base and Gemma-3-12B-Base on all 15 benchmarks. 
\item Similarly, Qwen3-14B-Base also achieves very competitive results compared to Qwen2.5-32B-Base with less than half of the parameters.
\item With only 1/5 activated non-embedding parameters, Qwen3-30B-A3B significantly outperforms Qwen2.5-14B-Base on all tasks, and achieves comparable performance to Qwen3-14B-Base and Qwen2.5-32B-Base, which brings us significant advantages in inference and training costs.
\end{enumerate}

\paragraph{Qwen3-8B / 4B / 1.7B / 0.6B-Base}
For edge-side models, we take similar-sized Qwen2.5, Llama-3, and Gemma-3 base models as the baselines. The results can be seen in Table~\ref{tab:base-8B}, Table~\ref{tab:base-4B}, and Table~\ref{tab:base-1B}. All Qwen3 8B / 4B / 1.7B / 0.6B-Base models continue to maintain strong performance across nearly all benchmarks. Notably, Qwen3-8B / 4B / 1.7B-Base models even outperform larger size Qwen2.5-14B / 7B / 3B Base models on over half of the benchmarks, especially on STEM-related and coding benchmarks, reflecting the significant improvement of the Qwen3 models.

\section{Post-training}

\label{sec:post}

\begin{figure}[htbp]
    \centering
    \includegraphics[width=\textwidth]{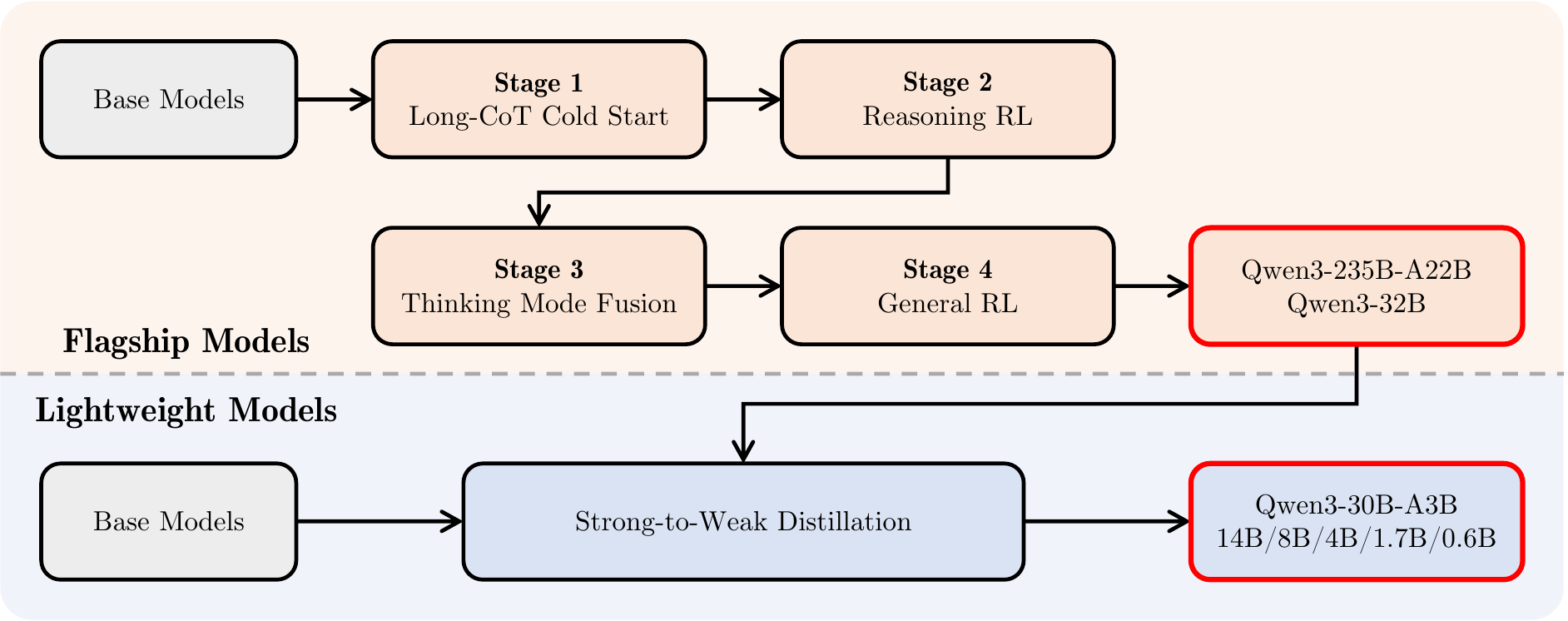}
    \caption{Post-training pipeline of the Qwen3 series models.}
    \vspace{1mm}
    \label{fig:post_training}
\end{figure}

The post-training pipeline of Qwen3 is strategically designed with two core objectives:

\begin{enumerate}[label=(\arabic*)]
    \item \textbf{Thinking Control}: 
    This involves the integration of two distinct modes, namely the ``non-thinking'' and ``thinking'' modes, providing users with the flexibility to choose whether the model should engage in reasoning or not, and to control the depth of thinking by specifying a token budget for the thinking process.
    \item \textbf{Strong-to-Weak Distillation}:
    This aims to streamline and optimize the post-training process for lightweight models.
    By leveraging the knowledge from large-scale models, we substantially reduce both the computational costs and the development efforts required for building smaller-scale models. 

\end{enumerate}

As illustrated in Figure~\ref{fig:post_training}, the flagship models in the Qwen3 series follow a sophisticated four-stage training process. The first two stages focus on developing the models' ``thinking'' abilities. The next two stages aim to integrate strong ``non-thinking'' functionalities into the models.

Preliminary experiments suggest that directly distilling the output logits from teacher models into lightweight student models can effectively enhance their performance while maintaining fine-grained control over their reasoning processes.
This approach eliminates the necessity of performing an exhaustive four-stage training process individually for every small-scale model.
It leads to better immediate performance, as indicated by higher Pass@1 scores, and also improves the model's ability of exploration, as reflected in improved Pass@64 results. 
In addition, it achieves these gains with much greater training efficiency, requiring only 1/10 of the GPU hours compared to the four-stage training method.

In the following sections, we present the four-stage training process and provide a detailed explanation of the Strong-to-Weak Distillation approach.

\subsection{Long-CoT Cold Start}

We begin by curating a comprehensive dataset that spans a wide range of categories, including math, code, logical reasoning, and general STEM problems. Each problem in the dataset is paired with verified reference answers or code-based test cases. This dataset serves as the foundation for the ``cold start'' phase of long Chain-of-Thought (long-CoT) training.

The dataset construction involves a rigorous two-phase filtering process: query filtering and response filtering.
In the query filtering phase, we use Qwen2.5-72B-Instruct to identify and remove queries that are not easily verifiable. This includes queries containing multiple sub-questions or those asking for general text generation. 
Furthermore, we exclude queries that Qwen2.5-72B-Instruct can answer correctly without using CoT reasoning. This helps prevent the model from relying on superficial guessing and ensures that only complex problems requiring deeper reasoning are included.
Additionally, we annotate each query's domain using Qwen2.5-72B-Instruct to maintain balanced domain representation across the dataset.

After reserving a validation query set, we generate $N$ candidate responses for each remaining query using QwQ-32B \citep{qwq32b}. 
When QwQ-32B consistently fails to generate correct solutions, human annotators manually assess the accuracy of the responses.
For queries with positive Pass@$N$, further stringent filtering criteria are applied to remove responses that (1) yield incorrect final answers, (2) contain substantial repetition, (3) clearly indicate guesswork without adequate reasoning, (4) exhibit inconsistencies between the thinking and summary contents, (5) involve inappropriate language mixing or stylistic shifts, or (6) are suspected of being overly similar to potential validation set items.
Subsequently, a carefully selected subset of the refined dataset is used for the initial cold-start training of the reasoning patterns. 
The objective at this stage is to instill foundational reasoning patterns in the model without overly emphasizing immediate reasoning performance. 
This approach ensures that the model's potential is not limited, allowing for greater flexibility and improvement during the subsequent reinforcement learning (RL) phase. 
To achieve this objective effectively, it is preferable to minimize both the number of training samples and the training steps during this preparatory phase.

\subsection{Reasoning RL}

The query-verifier pairs used in the Reasoning RL stage must satisfy the following four criteria:
(1) They were not used during the cold-start phase.
(2) They are learnable for the cold-start model.
(3) They are as challenging as possible.
(4) They cover a broad range of sub-domains.
We ultimately collect a total of 3,995 query-verifier pairs, and employed GRPO \citep{deepseekmath} to update the model parameters.
We observe that using a large batch size and a high number of rollouts per query, along with off-policy training to improve sample efficiency, is beneficial to the training process.
We have also addressed how to balance exploration and exploitation by controlling the model’s entropy to increase steadily or remain stable, which is crucial for maintaining stable training.
As a result, we achieve consistent improvements in both training reward and validation performance over the course of a single RL run, without any manual intervention on hyperparameters. For instance, the AIME'24 score of the Qwen3-235B-A22B model increases from 70.1 to 85.1 over a total of 170 RL training steps.

\subsection{Thinking Mode Fusion}

The goal of the Thinking Mode Fusion stage is to integrate the ``non-thinking'' capabilities into the previously developed ``thinking'' model. 
This approach allows developers to manage and control reasoning behaviors, while also reducing the cost and complexity of deploying separate models for thinking and non-thinking tasks.
To achieve this, we conduct continual supervised fine-tuning (SFT) on the Reasoning RL model and design a chat template to fuse the two modes. 
Moreover, we find that models capable of handling both modes proficiently perform consistently well under different thinking budgets.

\paragraph{Construction of SFT data.} The SFT dataset combines both the ``thinking'' and ``non-thinking'' data. 
To ensure that the performance of the Stage 2 model is not compromised by the additional SFT, the ``thinking'' data is generated via rejection sampling on Stage 1 queries using the Stage 2 model itself. 
The ``non-thinking'' data, on the other hand, is carefully curated to cover a diverse range of tasks, including coding, mathematics, instruction-following, multilingual tasks, creative writing, question answering, and role-playing.
Additionally, we employ automatically generated checklists for assessing the response quality of ``non-thinking'' data. To enhance the performance on tasks with low-resource languages, we particularly increase the proportion of translation tasks.

\paragraph{Chat Template Design.} To better integrate the two modes and enable users to dynamically switch the model's thinking process, we design chat templates for Qwen3, as shown in Table \ref{tab:thinking_fusion}. Specifically, for samples in thinking mode and non-thinking mode, we introduce \texttt{/think} and \texttt{/no\_think} flags in the user query or system message, respectively. This allows the model to follow the user's input and select the appropriate thinking mode accordingly.
For non-thinking mode samples, we retain an empty thinking block in the assistant's response. This design ensures internal format consistency within the model and allows developers to prevent the model from engaging in thinking behavior by concatenating an empty think block in the chat template. 
By default, the model operates in thinking mode; therefore, we add some thinking mode training samples where the user queries do not include \texttt{/think} flags. For more complex multi-turn dialogs, we randomly insert multiple \texttt{/think} and \texttt{/no\_think} flags into users' queries, with the model response adhering to the last flag encountered.

\paragraph{Thinking Budget.} An additional advantage of Thinking Mode Fusion is that, once the model learns to respond in both non-thinking and thinking modes, it naturally develops the ability to handle intermediate cases—generating responses based on incomplete thinking.
This capability lays the foundation for implementing budget control over the model's thinking process. Specifically, when the length of the model's thinking reaches a user-defined threshold, we manually halt the thinking process and insert the stop-thinking instruction: ``\texttt{{Considering the limited time by the user, I have to give the solution based on the thinking directly now.\textbackslash n</think>.\textbackslash n\textbackslash n}}''. 
After this instruction is inserted, the model proceeds to generate a final response based on its accumulated reasoning up to that point. It is worth noting that this ability is not explicitly trained but emerges naturally as a result of applying Thinking Mode Fusion.

\begin{table}[tbp]
\centering
\caption{\textbf{Examples of SFT data for thinking and non-thinking modes during the thinking mode fusion stage.} For the thinking mode, the \texttt{/think} flag can be omitted since it represents the default behavior.
This feature has been implemented in the chat template\footnote{\url{https://huggingface.co/Qwen/Qwen3-32B/blob/main/tokenizer_config.json}} supported by the Hugging Face's tokenizer, where the thinking mode can be disabled using an additional parameter \texttt{enable\_thinking=False}.
}
\vspace{-1mm}
\label{tab:thinking_fusion}
\adjustbox{center=\textwidth}{
\small
\begin{tabular}{@{}l|l@{}}
\toprule
\bf Thinking Mode & \bf Non-Thinking Mode \\
\midrule
\tabincell{l}{\texttt{<|im\_start|>user}\\
\texttt{\textcolor{blue}{\{query\}} \textcolor{red}{/think}<|im\_end|>}\\
\texttt{<|im\_start|>assistant}\\
\texttt{<think>}\\
\texttt{\textcolor{red}{\{thinking\_content\}}}\\
\texttt{</think>}\\
\\
\texttt{\textcolor{blue}{\{response\}}<|im\_end|>}\\
}
& 
\tabincell{l}{\texttt{<|im\_start|>user}\\
\texttt{\textcolor{blue}{\{query\}} \textcolor{red}{/no\_think}<|im\_end|>}\\
\texttt{<|im\_start|>assistant}\\
\texttt{<think>}\\
\\
\texttt{</think>}\\
\\
\texttt{\textcolor{blue}{\{response\}}<|im\_end|>}\\
}
\\
\bottomrule
\end{tabular}
}
\end{table}

\subsection{General RL}

The General RL stage aims to broadly enhance the models' capabilities and stability across diverse scenarios. To facilitate this, we have established a sophisticated \textbf{reward system} covering \textbf{over 20 distinct tasks}, each with customized scoring criteria. These tasks specifically target enhancements in the following core capabilities:
\begin{itemize}

\item \textbf{Instruction Following}: This capability ensures that models accurately interpret and follow user instructions, including requirements related to content, format, length, and the use of structured output, delivering responses that align with user expectations.

\item \textbf{Format Following}: In addition to explicit instructions, we expect the model to adhere to specific formatting conventions. For instance, it should respond appropriately to the \texttt{/think} and \texttt{/no\_think} flags by switching between thinking and non-thinking modes, and consistently use designated tokens (e.g., \texttt{<think>} and \texttt{</think>}) to separate the thinking and response parts in the final output.

\item \textbf{Preference Alignment}: For open-ended queries, preference alignment focuses on improving the model’s helpfulness, engagement, and style, ultimately delivering a more natural and satisfying user experience.

\item \textbf{Agent Ability}: This involves training the model to correctly invoke tools via designated interfaces. During the RL rollout, the model is allowed to perform complete multi-turn interaction cycles with real environment execution feedback, thereby improving its performance and stability in long-horizon decision-making tasks.

\item \textbf{Abilities for Specialized Scenarios}: In more specialized scenarios, we design tasks tailored to the specific context. For example, in Retrieval-Augmented Generation (RAG) tasks, we incorporate reward signals to guide the model toward generating accurate and contextually appropriate responses, thereby minimizing the risk of hallucination.

\end{itemize}

To provide feedback for the aforementioned tasks, we utilized three distinct types of rewards:

\begin{enumerate}[label=(\arabic*)]
    \item \textbf{Rule-based Reward}: The rule-based reward has been widely used in the reasoning RL stage, and is also useful for general tasks such as instruction following \citep{tulu3} and format adherence. Well-designed rule-based rewards can assess the correctness of model outputs with high precision, preventing issues like reward hacking.
    \item \textbf{Model-based Reward with Reference Answer}: In this approach, we provide a reference answer for each query and prompt Qwen2.5-72B-Instruct to score the model's response based on this reference. This method allows for more flexible handling of diverse tasks without requiring strict formatting, avoiding false negatives that can occur with purely rule-based rewards.
    \item \textbf{Model-based Reward without Reference Answer}: Leveraging human preference data, we train a reward model to assign scalar scores to model responses. This approach, which does not depend on a reference answer, can handle a broader range of queries while effectively enhancing the model's engagement and helpfulness.
\end{enumerate}

\subsection{Strong-to-Weak Distillation}

The Strong-to-Weak Distillation pipeline is specifically designed to optimize lightweight models, encompassing 5 dense models (Qwen3-0.6B, 1.7B, 4B, 8B, and 14B) and one MoE model (Qwen3-30B-A3B). This approach enhances model performance while effectively imparting robust mode-switching capabilities. The distillation process is divided into two primary phases:

\begin{enumerate}[label=(\arabic*)]
\item \textbf{Off-policy Distillation}: 
At this initial phase, we combine the outputs of teacher models generated with both \texttt{/think} and \texttt{/no\_think} modes for response distillation. This helps lightweight student models develop basic reasoning skills and the ability to switch between different modes of thinking, laying a solid foundation for the next on-policy training phase.

\item \textbf{On-policy Distillation}: 
In this phase, the student model generates on-policy sequences for fine-tuning. Specifically, prompts are sampled, and the student model produces responses in either \texttt{/think} or \texttt{/no\_think} mode. The student model is then fine-tuned by aligning its logits with those of a teacher model (Qwen3-32B or Qwen3-235B-A22B) to minimize the KL divergence.

\end{enumerate}

\subsection{Post-training Evaluation}

To comprehensively evaluate the quality of instruction-tuned models, we adopted automatic benchmarks to assess model performance under both thinking and non-thinking modes. These benchmarks are categorized into several dimensions:

\begin{itemize}
    \item \textbf{General Tasks}: We utilize benchmarks including MMLU-Redux \citep{mmluredux}, GPQA-Diamond \citep{gpqa}, C-Eval \citep{ceval}, and LiveBench~(2024-11-25) \citep{livebench}. For GPQA-Diamond, we sample 10 times for each query and report the averaged accuracy. 

    \item \textbf{Alignment Tasks}: To evaluate how well the model aligns with human preferences, we employ a suite of specialized benchmarks. For instruction-following performance, we report the strict-prompt accuracy of IFEval \citep{ifeval}. To assess alignment with human preferences on general topics, we utilize Arena-Hard \citep{arena-hard} and AlignBench v1.1 \citep{alignbench}. For writing tasks, we rely on Creative Writing V3 \citep{creative_writing} and WritingBench \citep{writingbench} to evaluate the model's proficiency and creativity.
    
    \item \textbf{Math \& Text Reasoning}: For evaluating mathematical and logical reasoning skills, we employ high-level math benchmarks including MATH-500 \citep{lightman2023lets}, AIME'24 and AIME'25 \citep{aime}, and text reasoning tasks including ZebraLogic \citep{zebralogic} and AutoLogi \citep{autologi}. For AIME problems, each year's questions include Part I and Part II, totaling 30 questions. For each question, we sample 64 times and take the average accuracy as the final score.
    
    \item \textbf{Agent \& Coding}: To test the model's proficiency in coding and agent-based tasks, we use BFCL v3 \citep{bfcl}, LiveCodeBench~(v5, 2024.10-2025.02) \citep{livecodebench}, and Codeforces Ratings from CodeElo \citep{codelo}. For BFCL, all Qwen3 models are evaluated using the FC format, and yarn was used to deploy the models to a context length of 64k for Multi-Turn evaluation. Some baselines are derived from the BFCL leaderboard, taking the higher scores between FC and Prompt formats. For models not reported on the leaderboard, the Prompt formats are evaluated. For LiveCodeBench, for the non-thinking mode, we use the officially recommended prompt, while for the thinking mode, we adjust the prompt template to allow the model to think more freely, by removing the restriction \texttt{You will not return anything except for the program}. To evaluate the performance gap between models and competitive programming experts, we use CodeForces to calculate Elo ratings. In our benchmark, each problem is solved by generating up to eight independent reasoning attempts.
    
    \item \textbf{Multilingual Tasks}: For multilingual capabilities, we evaluate four kinds of tasks: instruction following, knowledge, mathematics, and logical reasoning. Instruction following is assessed using Multi-IF \citep{he2024multi}, which focuses on 8 key languages. Knowledge assessment consisted of two types: regional knowledge evaluated through INCLUDE \citep{romanou2024includeevaluatingmultilinguallanguage}, covering 44 languages, and general knowledge assessed with MMMLU \citep{mmmlu} across 14 languages, excluding the unoptimized Yoruba language; for these two benchmarks, we sample only 10\% of the original data to improve evaluation efficiency. The mathematics task employ MT-AIME2024 \citep{son2025linguisticgeneralizabilitytesttimescaling}, encompassing 55 languages, and PolyMath \citep{wang2025polymathevaluatingmathematicalreasoning}, which includes 18 languages.  Logical reasoning is evaluated using MlogiQA, covering 10 languages, sourced from \citet{pmmeval}.
\end{itemize}

\begin{table}[tbp]
\centering
\caption{\textbf{Multilingual benchmarks and the included languages.} The languages are identified in IETF language tags.}
\vspace{-1mm}
\label{tab:multilingual-benchmark}
\begin{tabular}{lcl}
\toprule
Benchmark   & \# Langs & Languages \\
\midrule
Multi-IF    & 8 & \tabincell{l}{en, es, fr, hi, it, pt, ru, zh}   \\
INCLUDE     & 44 & \multirow{2}{*}{\tabincell{l}{ar, az, be, bg, bn, de, el, es, et, eu, fa, fi, fr, he, hi, hr, hu, hy, id, it, ja, ka, \\kk, ko, lt, mk, ml, ms, ne, nl, pl, pt, ru, sq, sr, ta, te, tl, tr, uk, ur, uz, vi, zh}}  \\
\\
MMMLU       & 14 & \tabincell{l}{ar, bn, de, en, es, fr, hi, id, it, ja, ko, pt, sw, zh} \\
MT-AIME2024 & 55 & \multirow{3}{*}{\tabincell{l}{af, ar, bg, bn, ca, cs, cy, da, de, el, en, es, et, fa, fi, fr, gu, he, hi, hr, hu, id, \\it, ja, kn, ko, lt, lv, mk, ml, mr, ne, nl, no, pa, pl, pt, ro, ru, sk, sl, so, sq, sv, \\ sw, ta, te, th, tl, tr, uk, ur, vi, zh-Hans, zh-Hant}}  \\
\\
\\
PolyMath    & 18 & \tabincell{l}{ar, bn, de, en, es, fr, id, it, ja, ko, ms, pt, ru, sw, te, th, vi, zh} \\
MLogiQA     & 10 & \tabincell{l}{ar, en, es, fr, ja, ko, pt, th, vi, zh} \\
\bottomrule
\end{tabular}
\end{table}

\begin{table}[tbp]
\centering
\caption{\textbf{Comparison among Qwen3-235B-A22B (Thinking) and other reasoning baselines. The highest and second-best scores are shown in \textbf{bold} and \underline{underlined}, respectively.}}
\vspace{-1mm}
\label{tab:instruct-235A22}
\adjustbox{center=\textwidth}{
\small
\setlength{\tabcolsep}{3pt} 
\begin{tabular}{@{}clccccc@{}}
\toprule
&  & \textbf{OpenAI-o1} & \textbf{DeepSeek-R1} & \tabincell{c}{\textbf{Grok-3-Beta}\\\textbf{(Think)}} & \textbf{Gemini2.5-Pro} & \textbf{Qwen3-235B-A22B} \\
\midrule
& Architecture & - & MoE & - & - & MoE \\
& \# Activated Params & - & 37B & - & - & 22B \\
& \# Total Params & - & 671B & - & - & 235B \\
\midrule
\multirow{4}{*}{\tabincell{c}{\textit{General}\\\textit{Tasks}}} & MMLU-Redux & 92.8 & \underline{92.9} & - & \textbf{93.7} & 92.7 \\
& GPQA-Diamond & 78.0 & 71.5 & \underline{80.2} & \textbf{84.0} & 71.1 \\
& C-Eval & 85.5 & \textbf{91.8} & - & 82.9 & \underline{89.6} \\
& LiveBench {\tiny{2024-11-25}} & 75.7 & 71.6 & - & \textbf{82.4} & \underline{77.1} \\
\midrule
\multirow{5}{*}{\tabincell{c}{\textit{Alignment}\\\textit{Tasks}}} & IFEval {\tiny{strict prompt}} & \textbf{92.6} & 83.3 & - & \underline{89.5} & 83.4 \\
& Arena-Hard & 92.1 & 92.3 & - & \textbf{96.4} & \underline{95.6} \\
& AlignBench v1.1 & 8.86 & 8.76 & - & \textbf{9.03} & \underline{8.94} \\
& Creative Writing v3 & 81.7 & \underline{85.5} & - & \textbf{86.0} & 84.6 \\
& WritingBench & 7.69 & 7.71 & - & \textbf{8.09} & \underline{8.03} \\
\midrule
\multirow{5}{*}{\tabincell{c}{\textit{Math \& Text}\\\textit{Reasoning}}} & MATH-500 & 96.4 & 97.3 &  & \textbf{98.8} & \underline{98.0} \\
& AIME'24 & 74.3 & 79.8 & 83.9 & \textbf{92.0} & \underline{85.7} \\
& AIME'25 & 79.2 & 70.0 & 77.3 & \textbf{86.7} & \underline{81.5} \\
& ZebraLogic & \underline{81.0} & 78.7 & - & \textbf{87.4} & 80.3 \\
& AutoLogi & 79.8 & \underline{86.1} & - & 85.4 & \textbf{89.0} \\
\midrule
\multirow{3}{*}{\tabincell{c}{\textit{Agent \&}\\\textit{Coding}}} & BFCL v3 & \underline{67.8} & 56.9 & - & 62.9 & \textbf{70.8} \\
& LiveCodeBench v5 & 63.9 & 64.3 & \underline{70.6} & 70.4 & \textbf{70.7} \\
& CodeForces {\tiny{(Rating / Percentile)}} & 1891 / 96.7\% & \underline{2029 / 98.1\%} & - & 2001 / 97.9\% & \textbf{2056 / 98.2\%} \\
\midrule
\multirow{6}{*}{\tabincell{c}{\textit{Multilingual}\\\textit{Tasks}}} & Multi-IF & 48.8 & 67.7 & - & \textbf{77.8} & \underline{71.9} \\
& INCLUDE & \underline{84.6} & 82.7 & - & \textbf{85.1} & 78.7 \\
& MMMLU {\tiny{14 languages}} & \textbf{88.4} & 86.4 & - & \underline{86.9} & 84.3 \\
& MT-AIME2024 & 67.4 & 73.5 & - & \underline{76.9} & \textbf{80.8} \\
& PolyMath & 38.9 & 47.1 & - & \underline{52.2} & \textbf{54.7} \\
& MLogiQA & 75.5 & 73.8 & - & \underline{75.6} & \textbf{77.1} \\
\bottomrule
\end{tabular}
}
\end{table}

\begin{table}[tbp]
\centering
\caption{\textbf{Comparison among Qwen3-235B-A22B (Non-thinking) and other non-reasoning baselines. The highest and second-best scores are shown in \textbf{bold} and \underline{underlined}, respectively.}}
\vspace{-1mm}
\label{tab:instruct-235A22-nothink}
\adjustbox{center=\textwidth}{
\small
\setlength{\tabcolsep}{3pt} 
\begin{tabular}{@{}clccccc@{}}
\toprule
&  & \tabincell{c}{\textbf{GPT-4o}\\ \textbf{-2024-11-20}} & \textbf{DeepSeek-V3} & \tabincell{c}{\textbf{Qwen2.5-72B}\\ \textbf{-Instruct}} & \tabincell{c}{\textbf{LLaMA-4}\\\textbf{-Maverick}} & \textbf{Qwen3-235B-A22B} \\
\midrule
& Architecture & - & MoE & Dense & MoE & MoE \\
& \# Activated Params & -  & 37B & 72B & 17B & 22B \\
& \# Total Params &  - & 671B & 72B & 402B & 235B \\
\midrule
\multirow{4}{*}{\tabincell{c}{\textit{General}\\\textit{Tasks}}}  & MMLU-Redux & 87.0 & 89.1 & 86.8 & \textbf{91.8} & \underline{89.2} \\
& GPQA-Diamond & 46.0 & 59.1 & 49.0 & \textbf{69.8} & \underline{62.9} \\
& C-Eval & 75.5 & \textbf{86.5} & 84.7 & 83.5 & \underline{86.1} \\
& LiveBench {\tiny{2024-11-25}} & 52.2 & \underline{60.5} & 51.4 & 59.5 & \textbf{62.5} \\
\midrule
\multirow{5}{*}{\tabincell{c}{\textit{Alignment}\\\textit{Tasks}}} & IFEval {\tiny{strict prompt}} & \underline{86.5} & 86.1 & 84.1 & \textbf{86.7} & 83.2 \\
& Arena-Hard & 85.3 & \underline{85.5} & 81.2 & 82.7 & \textbf{96.1} \\
& AlignBench v1.1 & 8.42 & \underline{8.64} & 7.89 & 7.97 & \textbf{8.91} \\
& Creative Writing v3 & \textbf{81.1} & 74.0 & 61.8 & 61.3 & \underline{80.4} \\
& WritingBench & \underline{7.11} & 6.49 & 7.06 & 5.46 & \textbf{7.70} \\
\midrule
\multirow{5}{*}{\tabincell{c}{\textit{Math \& Text}\\\textit{Reasoning}}} & MATH-500 & 77.2 & 90.2 & 83.6 & \underline{90.6} & \textbf{91.2} \\
& AIME'24 & 11.1 & \underline{39.2} & 18.9 & 38.5 & \textbf{40.1} \\
& AIME'25 & 7.6 & \textbf{28.8} & 15.0 & 15.9 & \underline{24.7} \\
& ZebraLogic & 27.4 & \textbf{42.1} & 26.6 & \underline{40.0} & 37.7 \\
& AutoLogi & 65.9 & \underline{76.1} & 66.1 & 75.2 & \textbf{83.3} \\
\midrule
\multirow{3}{*}{\tabincell{c}{\textit{Agent \&}\\\textit{Coding}}} & BFCL v3 & \textbf{72.5} & 57.6 & 63.4 & 52.9 & \underline{68.0} \\
& LiveCodeBench v5 & 32.7 & 33.1 & 30.7 & \textbf{37.2} & \underline{35.3} \\
& CodeForces {\tiny{(Rating / Percentile)}} & 864 / 35.4\% & \underline{1134 / 54.1\%} & 859 / 35.0\% & 712 / 24.3\% & \textbf{1387 / 75.7\%} \\
\midrule
\multirow{6}{*}{\tabincell{c}{\textit{Multilingual}\\\textit{Tasks}}} & Multi-IF & 65.6 & 55.6 & 65.3 & \textbf{75.5} & \underline{70.2} \\
& INCLUDE & \underline{78.8} & 76.7 & 69.6 & \textbf{80.9} & 75.6 \\
& MMMLU {\tiny{14 languages}} & 80.3 & \underline{81.1} & 76.9 & \textbf{82.5} & 79.8 \\
& MT-AIME2024 & 9.2 & 20.9 & 12.7 & \underline{27.0} & \textbf{32.4} \\
& PolyMath & 13.7 & 20.4 & 16.9 & \underline{26.1} & \textbf{27.0} \\
& MLogiQA & 57.4 & 58.9 & 59.3 & \underline{59.9} & \textbf{67.6} \\
\bottomrule
\end{tabular}
}
\end{table}

For all Qwen3 models in the thinking mode, we utilize a sampling temperature of 0.6, a top-p value of 0.95, and a top-k value of 20. Additionally, for Creative Writing v3 and WritingBench, we apply a presence penalty of 1.5 to encourage the generation of more diverse content. For Qwen3 models in the non-thinking mode, we configure the sampling hyperparameters with temperature = 0.7, top-p = 0.8, top-k = 20, and presence penalty = 1.5. For both the thinking and non-thinking modes, we set the max output length to 32,768 tokens, except AIME'24 and AIME'25 where we extend this length to 38,912 tokens to provide sufficient thinking space.

\paragraph{Summary of Evaluation Results}
From the evaluation results, we summarize several key conclusions of the finalized Qwen3 models as follows:
\begin{enumerate}[label=(\arabic*)]
    \item Our flagship model, Qwen3-235B-A22B, demonstrates the state-of-the-art overall performance among open-source models in both the thinking and non-thinking modes, surpassing strong baselines such as DeepSeek-R1 and DeepSeek-V3.
    Qwen3-235B-A22B is also highly competitive to closed-source leading models, such as OpenAI-o1, Gemini2.5-Pro, and GPT-4o, showcasing its profound reasoning capabilities and comprehensive general abilities.
    \item Our flagship dense model, Qwen3-32B, outperforms our previous strongest reasoning model, QwQ-32B, in most of the benchmarks, and performs comparably to the closed-source OpenAI-o3-mini, indicating its compelling reasoning capabilities.
    Qwen3-32B is also remarkably performant in the non-thinking mode and surpasses our previous flagship non-reasoning dense model, Qwen2.5-72B-Instruct.
    \item Our lightweight models, including Qwen3-30B-A3B, Qwen3-14B, and other smaller dense ones, possess consistently superior performance to the open-source models with a close or larger amount of parameters, proving the success of our Strong-to-Weak Distillation approach.
\end{enumerate}

The detailed results are as follows.

\paragraph{Qwen3-235B-A22B}
For our flagship model Qwen3-235B-A22B, we compare it with the leading reasoning and non-reasoning models.
For the thinking mode, we take OpenAI-o1 \citep{o1}, DeepSeek-R1 \citep{r1}, Grok-3-Beta (Think) \citep{grok3}, and Gemini2.5-Pro \citep{gemini2.5} as the reasoning baselines.
For the non-thinking mode, we take GPT-4o-2024-11-20 \citep{gpt4o}, DeepSeek-V3 \citep{deepseekv3}, Qwen2.5-72B-Instruct \citep{qwen2.5}, and LLaMA-4-Maverick \citep{llama4} as the non-reasoning baselines.
We present the evaluation results in Table~\ref{tab:instruct-235A22} and~\ref{tab:instruct-235A22-nothink}.

\begin{enumerate}[label=(\arabic*)]
    \item From Table~\ref{tab:instruct-235A22}, with only 60\% activated and 35\% total parameters, Qwen3-235B-A22B (Thinking) outperforms DeepSeek-R1 on \textbf{17/23} the benchmarks, particularly on the reasoning-demanded tasks (e.g., mathematics, agent, and coding), demonstrating the state-of-the-art reasoning capabilities of Qwen3-235B-A22B among open-source models.
    Moreover, Qwen3-235B-A22B (Thinking) is also highly competitive to the closed-source OpenAI-o1, Grok-3-Beta (Think), and Gemini2.5-Pro, substantially narrowing the gap in the reasoning capabilities between open-source and close-source models.
    \item From Table~\ref{tab:instruct-235A22-nothink}, Qwen3-235B-A22B (Non-thinking) exceeds the other leading open-source models, including DeepSeek-V3, LLaMA-4-Maverick, and our previous flagship model Qwen2.5-72B-Instruct, and also surpasses the closed-source GPT-4o-2024-11-20 in \textbf{18/23} the benchmarks, indicating its inherent strong capabilities even when not enhanced with the deliberate thinking process.
\end{enumerate}

\paragraph{Qwen3-32B}
For our flagship dense model, Qwen3-32B, we take DeepSeek-R1-Distill-Llama-70B, OpenAI-o3-mini (medium), and our previous strongest reasoning model, QwQ-32B \citep{qwq32b}, as the baselines in the thinking mode.
We also take GPT-4o-mini-2024-07-18, LLaMA-4-Scout, and our previous flagship model, Qwen2.5-72B-Instruct, as the baselines in the non-thinking mode.
We present the evaluation results in Table~\ref{tab:instruct-32B} and~\ref{tab:instruct-32B-nothink}.

\begin{enumerate}[label=(\arabic*)]
    \item From Table~\ref{tab:instruct-32B}, Qwen3-32B (Thinking) outperforms QwQ-32B on \textbf{17/23} the benchmarks, making it the new state-of-the-art reasoning model at the sweet size of 32B.
    Moreover, Qwen3-32B (Thinking) also competes with the closed-source OpenAI-o3-mini (medium) with better alignment and multilingual performance.
    \item From Table~\ref{tab:instruct-32B-nothink}, Qwen3-32B (Non-thinking) exhibits superior performance to all the baselines on almost all the benchmarks.
    Particularly, Qwen3-32B (Non-thinking) performs on par with Qwen2.5-72B-Instruct on the general tasks with significant advantages on the alignment, multilingual, and reasoning-related tasks, again proving the fundamental improvements of Qwen3 over our previous Qwen2.5 series models.
\end{enumerate}

\begin{table}[tbp]
\centering
\caption{\textbf{Comparison among Qwen3-32B (Thinking) and other reasoning baselines. The highest and second-best scores are shown in \textbf{bold} and \underline{underlined}, respectively.}}
\vspace{-1mm}
\label{tab:instruct-32B}
\adjustbox{center=\textwidth}{
\small
\begin{tabular}{@{}clcccc@{}}
\toprule
&  & \tabincell{c}{\textbf{DeepSeek-R1} \\ \textbf{-Distill-Llama-70B}} &  \textbf{QwQ-32B} &  \tabincell{c}{\textbf{OpenAI-o3-mini} \\ (medium)} & \textbf{Qwen3-32B} \\
\midrule
& Architecture &  Dense & Dense & - & Dense \\
& \# Activated Params & 70B & 32B & - & 32B \\
& \# Total Params & 70B & 32B & - & 32B \\
\midrule
\multirow{4}{*}{\tabincell{c}{\textit{General}\\\textit{Tasks}}} & MMLU-Redux & 89.3 & \underline{90.0} & \underline{90.0} & \textbf{90.9} \\
& GPQA-Diamond & 65.2 & 65.6 & \textbf{76.8} & \underline{68.4} \\
& C-Eval & 71.8 & \textbf{88.4} & 75.1 & \underline{87.3} \\
& LiveBench {\tiny{2024-11-25}} & 54.5 & \underline{72.0} & 70.0 & \textbf{74.9} \\
\bottomrule
\end{tabular}
}
\vspace{1mm}
\end{table}

\begin{table}[tbp]
\centering
\caption{\textbf{Comparison among Qwen3-32B (Non-thinking) and other non-reasoning baselines. The highest and second-best scores are shown in \textbf{bold} and \underline{underlined}, respectively.}}
\vspace{-1mm}
\label{tab:instruct-32B-nothink}
\adjustbox{center=\textwidth}{
\small
\begin{tabular}{@{}clcccc@{}}
\toprule
&  & \tabincell{c}{\textbf{GPT-4o-mini}\\\textbf{-2024-07-18}} & \tabincell{c}{\textbf{LLaMA-4}\\\textbf{-Scout}} & \tabincell{c}{\textbf{Qwen2.5-72B}\\\textbf{-Instruct}} & \textbf{Qwen3-32B} \\
\midrule
& Architecture & - & MoE & Dense & Dense \\
& \# Activated Params & - & 17B & 72B & 32B \\
& \# Total Params & - & 109B & 72B & 32B \\
\midrule
\multirow{4}{*}{\tabincell{c}{\textit{General}\\\textit{Tasks}}} & MMLU-Redux & 81.5 & \underline{86.3} & \textbf{86.8} & 85.7 \\
& GPQA-Diamond & 40.2 & \textbf{57.2} & 49.0 & \underline{54.6} \\
& C-Eval & 66.3 & 78.2 & \textbf{84.7} & \underline{83.3} \\
& LiveBench {\tiny{2024-11-25}} & 41.3 & 47.6 & \underline{51.4} & \textbf{59.8} \\
\bottomrule
\end{tabular}
}
\end{table}

\begin{table}[tbp]
\centering
\caption{\textbf{Comparison among Qwen3-30B-A3B / Qwen3-14B (Thinking) and other reasoning baselines. The highest and second-best scores are shown in \textbf{bold} and \underline{underlined}, respectively.}}
\vspace{-1mm}
\label{tab:instruct-30A3}
\adjustbox{center=\textwidth}{
\small
\begin{tabular}{@{}clcccc@{}}
\toprule
&  & \tabincell{c}{\textbf{DeepSeek-R1}\\\textbf{-Distill-Qwen-32B}} & \textbf{QwQ-32B} & \textbf{Qwen3-14B} & \textbf{Qwen3-30B-A3B} \\
\midrule
& Architecture & Dense & Dense &Dense & MoE \\
& \# Activated Params & 32B & 32B  & 14B & 3B\\
& \# Total Params & 32B & 32B  & 14B & 30B \\
\midrule
\multirow{4}{*}{\tabincell{c}{\textit{General}\\\textit{Tasks}}} & MMLU-Redux & 88.2 & \textbf{90.0} & 88.6 & \underline{89.5} \\
& GPQA-Diamond & 62.1 & \underline{65.6} & 64.0 & \textbf{65.8} \\
& C-Eval & 82.2 & \textbf{88.4} & 86.2 & \underline{86.6} \\
& LiveBench {\tiny{2024-11-25}} & 45.6 & \underline{72.0} & 71.3 & \textbf{74.3} \\
\bottomrule
\end{tabular}
}
\vspace{1mm}
\end{table}

\begin{table}[tbp]
\centering
\caption{\textbf{Comparison among Qwen3-30B-A3B / Qwen3-14B (Non-thinking) and other non-reasoning baselines. The highest and second-best scores are shown in \textbf{bold} and \underline{underlined}, respectively.}}
\vspace{-1mm}
\label{tab:instruct-30A3-nothink}
\adjustbox{center=\textwidth}{
\small
\setlength{\tabcolsep}{3pt} %
\begin{tabular}{@{}clccccc@{}}
\toprule
& & \textbf{Phi-4} & \tabincell{c}{\textbf{Gemma-3}\\\textbf{-27B-IT}} & \tabincell{c}{\textbf{Qwen2.5-32B}\\\textbf{-Instruct}} & \textbf{Qwen3-14B} & \textbf{Qwen3-30B-A3B} \\
\midrule
& Architecture & Dense & Dense & Dense & Dense & MoE \\
& \# Activated Params & 14B & 27B & 32B & 14B & 3B \\
& \# Total Params & 14B & 27B & 32B & 14B & 30B \\
\midrule
\multirow{4}{*}{\tabincell{c}{\textit{General}\\\textit{Tasks}}} & MMLU-Redux & \textbf{85.3} & 82.6 & 83.9 & 82.0 & \underline{84.1} \\
& GPQA-Diamond & \textbf{56.1} & 42.4 & 49.5 & \underline{54.8} & \underline{54.8} \\
& C-Eval & 66.9 & 66.6 & 80.6 & \underline{81.0} & \textbf{82.9} \\
& LiveBench {\tiny{2024-11-25}} & 41.6 & 49.2 & 50.0 & \textbf{59.6} & \underline{59.4} \\
\bottomrule
\end{tabular}
}
\end{table}

\begin{table}[tbp]
\centering
\caption{\textbf{Comparison among Qwen3-8B / Qwen3-4B (Thinking) and other reasoning baselines. The highest and second-best scores are shown in \textbf{bold} and \underline{underlined}, respectively.}}
\vspace{-1mm}
\label{tab:instruct-8B}
\adjustbox{center=\textwidth}{
\small
\begin{tabular}{@{}clcccc@{}}
\toprule
& & \tabincell{c}{\textbf{DeepSeek-R1}\\\textbf{-Distill-Qwen-14B}} & \tabincell{c}{\textbf{DeepSeek-R1}\\\textbf{-Distill-Qwen-32B}} & \textbf{Qwen3-4B} & \textbf{Qwen3-8B} \\
\midrule
& Architecture & Dense & Dense & Dense & Dense \\
& \# Activated Params & 14B & 32B & 4B & 8B \\
& \# Total Params & 14B & 32B & 4B & 8B \\
\midrule
\multirow{4}{*}{\tabincell{c}{\textit{General}\\\textit{Tasks}}} & MMLU-Redux & 84.1 & \textbf{88.2} & 83.7 & \underline{87.5} \\
& GPQA-Diamond & 59.1 & \textbf{62.1} & 55.9 & \underline{62.0} \\
& C-Eval & 78.1 & \underline{82.2} & 77.5 & \textbf{83.4} \\
& LiveBench {\tiny{2024-11-25}} & 52.3 & 45.6 & \underline{63.6} & \textbf{67.1} \\
\bottomrule
\end{tabular}
}
\vspace{1mm}
\end{table}

\begin{table}[tbp]
\centering
\caption{\textbf{Comparison among Qwen3-8B / Qwen3-4B (Non-thinking) and other non-reasoning baselines. The highest and second-best scores are shown in \textbf{bold} and \underline{underlined}, respectively.}}
\vspace{-1mm}
\label{tab:instruct-8B-nothink}
\adjustbox{center=\textwidth}{
\small
\setlength{\tabcolsep}{3pt}
\begin{tabular}{@{}clcccccc@{}}
\toprule
& & \tabincell{c}{\textbf{LLaMA-3.1-8B}\\\textbf{-Instruct}} & \tabincell{c}{\textbf{Gemma-3}\\\textbf{-12B-IT}} & \tabincell{c}{\textbf{Qwen2.5-7B}\\\textbf{-Instruct}} & \tabincell{c}{\textbf{Qwen2.5-14B}\\\textbf{-Instruct}} & \textbf{Qwen3-4B} & \textbf{Qwen3-8B} \\
\midrule
& Architecture & Dense & Dense & Dense & Dense & Dense & Dense \\
& \# Activated Params & 8B & 12B & 7B & 14B & 4B & 8B\\
& \# Total Params & 8B & 12B & 7B & 14B & 4B & 8B\\
\midrule
\multirow{4}{*}{\tabincell{c}{\textit{General}\\\textit{Tasks}}} & MMLU-Redux & 61.7 & 77.8 & 75.4 & \textbf{80.0} & 77.3 & \underline{79.5} \\
& GPQA-Diamond & 32.8 & 40.9 & 36.4 & \textbf{45.5} & \underline{41.7} & 39.3 \\
& C-Eval & 52.0 & 61.1 & 76.2 & \textbf{78.0} & 72.2 & \underline{77.9} \\
& LiveBench {\tiny{2024-11-25}} & 26.0 & 43.7 & 34.9 & 42.2 & \underline{48.4} & \textbf{53.5} \\
\bottomrule
\end{tabular}
}
\end{table}

\begin{table}[tbp]
\centering
\caption{\textbf{Comparison among Qwen3-1.7B / Qwen3-0.6B (Thinking) and other reasoning baselines. The highest and second-best scores are shown in \textbf{bold} and \underline{underlined}, respectively.}}
\vspace{-1mm}
\label{tab:instruct-1.7B}
\adjustbox{center=\textwidth}{
\small
\begin{tabular}{@{}clcccc@{}}
\toprule
&  & \tabincell{c}{\textbf{DeepSeek-R1}\\\textbf{-Distill-Qwen-1.5B}} & \tabincell{c}{\textbf{DeepSeek-R1}\\\textbf{-Distill-Llama-8B}} & \textbf{Qwen3-0.6B} & \textbf{Qwen3-1.7B} \\
\midrule
& Architecture & Dense & Dense & Dense & Dense \\
& \# Activated Params & 1.5B & 8B & 0.6B & 1.7B \\
& \# Total Params & 1.5B & 8B & 0.6B & 1.7B \\
\midrule
\multirow{4}{*}{\tabincell{c}{\textit{General}\\\textit{Tasks}}} & MMLU-Redux & 45.4 & \underline{66.4} & 55.6 & \textbf{73.9} \\
& GPQA-Diamond & 33.8 & \textbf{49.0} & 27.9 & \underline{40.1} \\
& C-Eval & 27.1 & \underline{50.4} & \underline{50.4} & \textbf{68.1} \\
& LiveBench {\tiny{2024-11-25}} & 24.9 & \underline{40.6} & 30.3 & \textbf{51.1} \\
\bottomrule
\end{tabular}
}
\vspace{1mm}
\end{table}

\begin{table}[tbp]
\centering
\caption{\textbf{Comparison among Qwen3-1.7B / Qwen3-0.6B (Non-thinking) and other non-reasoning baselines. The highest and second-best scores are shown in \textbf{bold} and \underline{underlined}, respectively.}}
\vspace{-1mm}
\label{tab:instruct-1.7B-nothink}
\adjustbox{center=\textwidth}{
\small
\setlength{\tabcolsep}{3pt}
\begin{tabular}{@{}clccccccc@{}}
\toprule
 & & \tabincell{c}{\textbf{Gemma-3}\\\textbf{-1B-IT}} & \textbf{Phi-4-mini} & \tabincell{c}{\textbf{Qwen2.5-1.5B}\\\textbf{-Instruct}} & \tabincell{c}{\textbf{Qwen2.5-3B}\\\textbf{-Instruct}} & \textbf{Qwen3-0.6B} & \textbf{Qwen3-1.7B} \\
\midrule
& Architecture  & Dense & Dense & Dense & Dense & Dense & Dense \\
& \# Activated Params  & 1.0B & 3.8B & 1.5B & 3.1B & 0.6B & 1.7B\\
& \# Total Params & 1.0B & 3.8B & 1.5B & 3.1B & 0.6B & 1.7B\\
\midrule
\multirow{4}{*}{\tabincell{c}{\textit{General}\\\textit{Tasks}}} & MMLU-Redux & 33.3 & \textbf{67.9} & 50.7 & \underline{64.4} & 44.6 & \underline{64.4} \\
& GPQA-Diamond & 19.2 & 25.2 & \underline{29.8} & \textbf{30.3} & 22.9 & 28.6 \\
& C-Eval & 28.5 & 40.0 & 53.3 & \textbf{68.2} & 42.6 & \underline{61.0} \\
& LiveBench {\tiny{2024-11-25}} & 14.4 & \underline{25.3} & 18.0 & 23.8 & 21.8 & \textbf{35.6} \\
\bottomrule
\end{tabular}
}
\end{table}

\paragraph{Qwen3-30B-A3B \& Qwen3-14B}
For Qwen3-30B-A3B and Qwen3-14B, we compare them with DeepSeek-R1-Distill-Qwen-32B and QwQ-32B in the thinking mode, and Phi-4 \citep{phi4}, Gemma-3-27B-IT \citep{gemma3}, and Qwen2.5-32B-Instruct in the non-thinking mode, respectively.
We present the evaluation results in Table~\ref{tab:instruct-30A3} and~\ref{tab:instruct-30A3-nothink}.

\begin{enumerate}[label=(\arabic*)]
    \item From Table~\ref{tab:instruct-30A3}, Qwen3-30B-A3B and Qwen3-14B (Thinking) are both highly competitive to QwQ-32B, especially on the reasoning-related benchmarks.
    It is noteworthy that Qwen3-30B-A3B achieves comparable performance to QwQ-32B with a smaller model size and less than \textbf{1/10} activated parameters, demonstrating the effectiveness of our Strong-to-Weak Distillation approach in endowing lightweight models with profound reasoning capabilities.
    \item From Table~\ref{tab:instruct-30A3-nothink}, Qwen3-30B-A3B and Qwen3-14B (Non-thinking) surpass the non-reasoning baselines in most of the benchmarks.
    They exceed our previous Qwen2.5-32B-Instruct model with significantly fewer activated and total parameters, allowing for more efficient and cost-effective performance.
\end{enumerate}

\paragraph{Qwen3-8B / 4B / 1.7B / 0.6B}
For Qwen3-8B and Qwen3-4B, we compare them with DeepSeek-R1-Distill-Qwen-14B and DeepSeek-R1-Distill-Qwen-32B in the thinking mode, and LLaMA-3.1-8B-Instruct \citep{llama3}, Gemma-3-12B-IT \citep{gemma3}, Qwen2.5-7B-Instruct, and Qwen2.5-14B-Instruct in the non-thinking mode, respectively.
For Qwen3-1.7B and Qwen3-0.6B, we compare them with DeepSeek-R1-Distill-Qwen-1.5B and DeepSeek-R1-Distill-Llama-8B in the thinking mode, and Gemma-3-1B-IT, Phi-4-mini, Qwen2.5-1.5B-Instruct, and Qwen2.5-3B-Instruct in the non-thinking mode, respectively.
We present the evaluation results of Qwen3-8B and Qwen3-4B in Table~\ref{tab:instruct-8B} and~\ref{tab:instruct-8B-nothink} and those of Qwen3-1.7B and Qwen3-0.6B in Table~\ref{tab:instruct-1.7B} and~\ref{tab:instruct-1.7B-nothink}, respectively.
Overall, these edge-side models exhibit impressive performance and outperform baselines even with more parameters, including our previous Qwen2.5 models, in either the thinking or the non-thinking mode.
These results, once again, demonstrate the efficacy of our Strong-to-Weak Distillation approach, making it possible for us to build the lightweight Qwen3 models with remarkably reduced costs and efforts.

\subsection{Discussion}

\paragraph{The Effectiveness of Thinking Budget}

\begin{figure}[htbp]
    \centering
    \includegraphics[width=\textwidth]{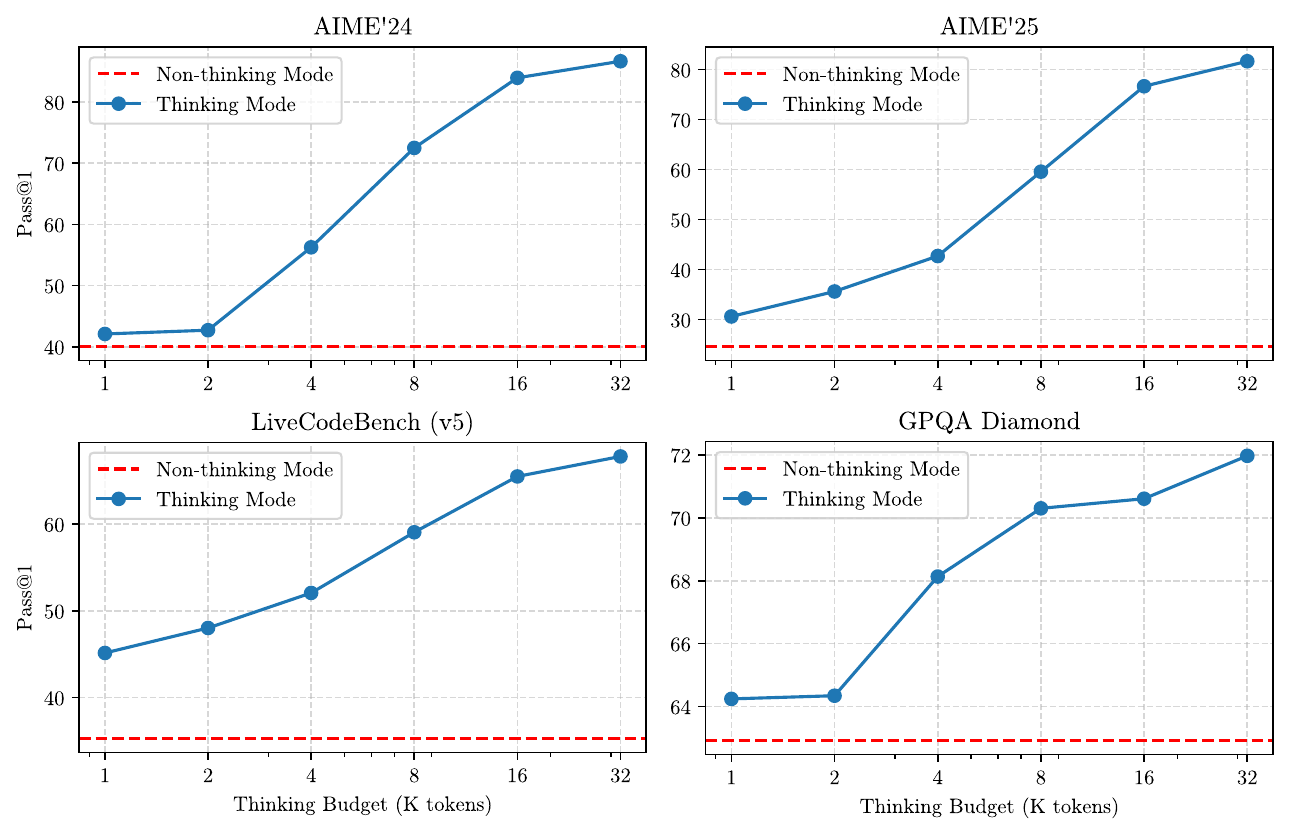}
    \caption{Performance of Qwen3-235B-A22B with respect to the thinking budget.}
    \label{fig:thinking_budget}
\end{figure}

To verify that Qwen3 can enhance its intelligence level by leveraging an increased thinking budget, we adjust the allocated thinking budget on four benchmarks across Mathematics, Coding, and STEM domains. The resulting scaling curves are presented in Figure~\ref{fig:thinking_budget}, Qwen3 demonstrates scalable and smooth performance improvements correlated to the allocated thinking budget.
Moreover, we observe that if we further extend the output length beyond 32K, the model's performance is expected to improve further in the future. We leave this exploration as future work.

\paragraph{The Effectiveness and Efficiency of On-Policy Distillation}

We evaluate the effectiveness and efficiency of on-policy distillation by comparing the performance and computational cost—measured in GPU hours—after undergoing distillation versus direct reinforcement learning, both starting from the same off-policy distilled 8B checkpoint. For simplicity, we focus solely on math and code-related queries in this comparison. The results, summarized in Table \ref{tab:distillation}, show that distillation achieves significantly better performance than reinforcement learning while requiring approximately only $1/10$ of the GPU hours.
Furthermore, distillation from teacher logits enables the student model to expand its exploration space and enhance its reasoning potential, as evidenced by the improved pass@64 scores on the AIME'24 and AIME'25 benchmarks after distillation, compared to the initial checkpoint. In contrast, reinforcement learning does not lead to any improvement in pass@64 scores.
These observations highlight the advantages of leveraging a stronger teacher model in guiding student model learning.

\begin{table}[htbp]
\caption{Comparison of reinforcement learning and on-policy distillation on Qwen3-8B. Numbers in parentheses indicate pass@64 scores.}
\small
\centering
\setlength\tabcolsep{3pt}
\begin{tabular}{@{}lcccccc|c@{}} 
\toprule
Method & AIME'24 & AIME'25 & MATH500 & \tabincell{c}{LiveCodeBench\\v5} & \tabincell{c}{MMLU\\-Redux} & \tabincell{c}{GPQA\\-Diamond} & \tabincell{c}{GPU\\Hours}\\
\midrule
Off-policy Distillation & 55.0 (90.0) & 42.8 (83.3) & 92.4 & 42.0 & 86.4 & 55.6 & - \\
 + Reinforcement Learning & 67.6 (90.0) & 55.5 (83.3) & 94.8 & 52.9 & 86.9 & 61.3 & 17,920 \\
+ On-policy Distillation & \textbf{74.4} (\textbf{93.3}) & \textbf{65.5} (\textbf{86.7}) & \textbf{97.0} & \textbf{60.3} &  \textbf{88.3}  & \textbf{63.3} & 1,800 \\
\bottomrule
\end{tabular}
\label{tab:distillation}
\end{table}

\paragraph{The Effects of Thinking Mode Fusion and General RL}

To evaluate the effectiveness of Thinking Mode Fusion and General Reinforcement Learning (RL) during the post-training, we conduct evaluations on various stages of the Qwen-32B model. In addition to the datasets mentioned earlier, we introduce several in-house benchmarks to monitor other capabilities. These benchmarks include:

\begin{itemize}
    \item \textbf{CounterFactQA}: Contains counterfactual questions where the model needs to identify that the questions are not factual and avoid generating hallucinatory answers.
    \item \textbf{LengthCtrl}: Includes creative writing tasks with length requirements; the final score is based on the difference between the generated content length and the target length.
    \item \textbf{ThinkFollow}: Involves multi-turn dialogues with randomly inserted \texttt{/think} and \texttt{/no\_think} flags to test whether the model can correctly switch thinking modes based on user queries.
    \item \textbf{ToolUse}: Evaluates the stability of the model in single-turn, multi-turn, and multi-step tool calling processes. The score includes accuracy in intent recognition, format accuracy, and parameter accuracy during the tool calling process.
\end{itemize}

\begin{table}[htbp]
\caption{Performance of Qwen3-32B after Reasoning RL (Stage 2), Thinking Mode Fusion (Stage 3), and General RL (Stage 4). Benchmarks with * are in-house datasets.}
\small
\centering
\begin{tabular}{@{}clccccc@{}}
\toprule
&  & \tabincell{c}{Stage 2\\Reasoning RL} & \multicolumn{2}{c}{\tabincell{c}{Stage 3\\Thinking Mode Fusion}} & \multicolumn{2}{c}{\tabincell{c}{Stage 4\\General RL}} \\
\cmidrule(r){3-3}\cmidrule(lr){4-5}\cmidrule(l){6-7}
&  Benchmark & \tiny Thinking & \tiny Thinking & \tiny Non-Thinking & \tiny Thinking & \tiny Non-Thinking \\
\midrule
\multirow{3}{*}{\tabincell{c}{\textit{General} \\ \textit{Tasks}}} & 
LiveBench {\tiny 2024-11-25} & 68.6 & 70.9{\tiny \textcolor{ForestGreen}{+2.3}} & 57.1 & 74.9{\tiny \textcolor{ForestGreen}{+4.0}} & 59.8{\tiny \textcolor{ForestGreen}{+2.8}} \\
& Arena-Hard & 86.8 & 89.4{\tiny \textcolor{ForestGreen}{+2.6}} & 88.5 & 93.8{\tiny \textcolor{ForestGreen}{+4.4}} & 92.8{\tiny \textcolor{ForestGreen}{+4.3}} \\
& CounterFactQA* & 50.4 & 61.3{\tiny \textcolor{ForestGreen}{+10.9}} & 64.3 & 68.1{\tiny \textcolor{ForestGreen}{+6.8}} & 66.4{\tiny \textcolor{ForestGreen}{+2.1}} \\
\midrule
\multirow{4}{*}{\tabincell{c}{\textit{Instruction}\\\textit{\& Format}\\\textit{Following}}} & IFEval {\tiny strict prompt} & 73.0 & 78.4{\tiny \textcolor{ForestGreen}{+5.4}} & 78.4 & 85.0{\tiny \textcolor{ForestGreen}{+6.6}} & 83.2{\tiny \textcolor{ForestGreen}{+4.8}} \\
& Multi-IF & 61.4 & 64.6{\tiny \textcolor{ForestGreen}{+3.2}} & 65.2 & 73.0{\tiny \textcolor{ForestGreen}{+8.4}} & 70.7{\tiny \textcolor{ForestGreen}{+5.5}} \\
& LengthCtrl* & 62.6 & 70.6{\tiny \textcolor{ForestGreen}{+8.0}} & 84.9 & 73.5{\tiny \textcolor{ForestGreen}{+2.9}} & 87.3{\tiny \textcolor{ForestGreen}{+2.4}} \\
& ThinkFollow* & - & \multicolumn{2}{c}{88.7} & \multicolumn{2}{c}{98.9{\tiny \textcolor{ForestGreen}{+10.2}}} \\
\midrule
\multirow{2}{*}{\tabincell{c}{\textit{Agent}}}
& BFCL v3 & 69.0 & 68.4{\tiny \textcolor{OrangeRed}{-0.6}} & 61.5 & 70.3{\tiny \textcolor{ForestGreen}{+1.9}} & 63.0{\tiny \textcolor{ForestGreen}{+1.5}} \\
& ToolUse* & 63.3 & 70.4{\tiny \textcolor{ForestGreen}{+7.1}} & 73.2 & 85.5{\tiny \textcolor{ForestGreen}{+15.1}} & 86.5{\tiny \textcolor{ForestGreen}{+13.3}} \\
\midrule
\multirow{2}{*}{\tabincell{c}{\textit{Knowledge \&} \\ \textit{STEM}}}
& MMLU-Redux & 91.4 & 91.0{\tiny \textcolor{OrangeRed}{-0.4}} & 86.7 & 90.9{\tiny \textcolor{OrangeRed}{-0.1}} & 85.7{\tiny \textcolor{OrangeRed}{-1.0}} \\
& GPQA-Diamond & 68.8 & 69.0{\tiny \textcolor{ForestGreen}{+0.2}} & 50.4 & 68.4{\tiny \textcolor{OrangeRed}{-0.6}} & 54.6{\tiny \textcolor{ForestGreen}{+4.3}} \\
\midrule
\multirow{2}{*}{\tabincell{c}{\textit{Math \&} \\ \textit{Coding}}}
& AIME'24 & 83.8 & 81.9{\tiny \textcolor{OrangeRed}{-1.9}} & 28.5 & 81.4{\tiny \textcolor{OrangeRed}{-0.5}} & 31.0{\tiny \textcolor{ForestGreen}{+2.5}} \\
& LiveCodeBench v5 & 68.4 & 67.2{\tiny \textcolor{OrangeRed}{-1.2}} & 31.1 & 65.7{\tiny \textcolor{OrangeRed}{-1.5}} & 31.3{\tiny \textcolor{ForestGreen}{+0.2}} \\
\bottomrule
\end{tabular}
\label{tab:stage_ablation}
\vspace{1mm}
\end{table}

The results are shown in Table~\ref{tab:stage_ablation}, where we can draw the following conclusions:

\begin{enumerate}[label=(\arabic*)]
    \item Stage 3 integrates the non-thinking mode into the model, which already possesses thinking capabilities after the first two stages of training. The ThinkFollow benchmark score of 88.7 indicates that the model has developed an initial ability to switch between modes, though it still occasionally makes errors. Stage 3 also enhances the model's general and instruction-following capabilities in thinking mode, with CounterFactQA improving by 10.9 points and LengthCtrl by 8.0 points.
    
    \item Stage 4 further strengthens the model's general, instruction-following, and agent capabilities in both thinking and non-thinking modes. Notably, the ThinkFollow score improves to 98.9, ensuring accurate mode switching.
    
    \item For Knowledge, STEM, Math, and Coding tasks, Thinking Mode Fusion and General RL do not bring significant improvements. In contrast, for challenging tasks like AIME'24 and LiveCodeBench, the performance in thinking mode actually decreases after these two training stages. We conjecture this degradation is due to the model being trained on a broader range of general tasks, which may compromise its specialized capabilities in handling complex problems. During the development of Qwen3, we choose to accept this performance trade-off to enhance the model's overall versatility.
\end{enumerate}


\section{Conclusion}

In this technical report, we introduce Qwen3, the latest version of the Qwen series. Qwen3 features both thinking mode and non-thinking mode, allowing users to dynamically manage the number of tokens used for complex thinking tasks. The model was pre-trained on an extensive dataset containing 36 trillion tokens, enabling it to understand and generate text in 119 languages and dialects. Through a series of comprehensive evaluations, Qwen3 has shown strong performance across a range of standard benchmarks for both pre-trained and post-trained models, including tasks related to code generation, mathematics, reasoning, and agents.

In the near future, our research will focus on several key areas. We will continue to scale up pretraining by using data that is both higher in quality and more diverse in content. At the same time, we will work on improving model architecture and training methods for the purposes of effective compression, scaling to extremely long contexts, etc.
In addition, we plan to increase computational resources for reinforcement learning, with a particular emphasis on agent-based RL systems that learn from environmental feedback. This will allow us to build agents capable of tackling complex tasks that require inference time scaling.

\section{Authors}

\textbf{Core Contributors:} An Yang, Anfeng Li, Baosong Yang, Beichen Zhang, Binyuan Hui, Bo Zheng, Bowen Yu, Chang Gao, Chengen Huang, Chenxu Lv, Chujie Zheng, Dayiheng Liu, Fan Zhou, Fei Huang, Feng Hu, Hao Ge, Haoran Wei, Huan Lin, Jialong Tang, Jian Yang, Jianhong Tu, Jianwei Zhang, Jianxin Yang, Jiaxi Yang, Jing Zhou, Jingren Zhou, Junyang Lin, Kai Dang, Keqin Bao, Kexin Yang, Le Yu, Lianghao Deng, Mei Li, Mingfeng Xue, Mingze Li, Pei Zhang, Peng Wang, Qin Zhu, Rui Men, Ruize Gao, Shixuan Liu, Shuang Luo, Tianhao Li, Tianyi Tang, Wenbiao Yin, Xingzhang Ren, Xinyu Wang, Xinyu Zhang, Xuancheng Ren, Yang Fan, Yang Su, Yichang Zhang, Yinger Zhang, Yu Wan, Yuqiong Liu, Zekun Wang, Zeyu Cui, Zhenru Zhang, Zhipeng Zhou, Zihan Qiu

\textbf{Contributors:} Bei Chen, Biao Sun, Bin Luo, Bin Zhang, Binghai Wang, Bowen Ping, Boyi Deng, Chang Si, Chaojie Yang, Chen Cheng, Chenfei Wu, Chengpeng Li, Chengyuan Li, Fan Hong, Guobin Zhao, Hang Zhang, Hangrui Hu, Hanyu Zhao, Hao Lin, Hao Xiang, Haoyan Huang, Hongkun Hao, Humen Zhong, Jialin Wang, Jiandong Jiang, Jianqiang Wan, Jianyuan Zeng, Jiawei Chen, Jie Zhang, Jin Xu, Jinkai Wang, Jinyang Zhang, Jinzheng He, Jun Tang, Kai Zhang, Ke Yi, Keming Lu, Keqin Chen, Langshi Chen, Le Jiang, Lei Zhang, Linjuan Wu, Man Yuan, Mingkun Yang, Minmin Sun, Mouxiang Chen, Na Ni, Nuo Chen, Peng Liu, Peng Wang, Peng Zhu, Pengcheng Zhang, Pengfei Wang, Qiaoyu Tang, Qing Fu, Qiuyue Wang, Rong Zhang, Rui Hu, Runji Lin, Shen Huang, Shuai Bai, Shutong Jiang, Sibo Song, Siqi Zhang, Song Chen, Tao He, Ting He, Tingfeng Hui, Wei Ding, Wei Liao, Wei Lin, Wei Zhang, Weijia Xu, Wenbin Ge, Wenmeng Zhou, Wenyuan Yu, Xianyan Jia, Xianzhong Shi, Xiaodong Deng, Xiaoming Huang, Xiaoyuan Li, Ximing Zhou, Xinyao Niu, Xipin Wei, Xuejing Liu, Yang Liu, Yang Yao, Yang Zhang, Yanpeng Li, Yantao Liu, Yidan Zhang, Yikai Zhu, Yiming Wang, Yiwen Hu, Yong Jiang, Yong Li, Yongan Yue, Yu Guan, Yuanzhi Zhu, Yunfei Chu, Yunlong Feng, Yuxin Zhou, Yuxuan Cai, Zeyao Ma, Zhaohai Li, Zheng Li, Zhengyang Tang, Zheren Fu, Zhi Li, Zhibo Yang, Zhifang Guo, Zhipeng Zhang, Zhiying Xu, Zhiyu Yin, Zhongshen Zeng, Zile Qiao, Ziye Meng, Zongmeng Zhang

\clearpage
\appendix

\section{Appendix}
\label{sec:appendix}

\subsection{Additional Evaluation Results}
\subsubsection{Long-Context Ability}

\begin{table}[h]
\centering
\caption{\textbf{Performance of Qwen3 Models on the RULER benchmark.}}
\label{tab:ruler}
\small
\begin{tabular}{@{}cllllllll@{}}
\toprule
& \multirow{2}[2]{*}{\bf Model} & \multicolumn{6}{c}{\bf RULER}  \\ 
\cmidrule{3-9}
&  &  \bf Avg.  & \bf 4K   & \bf 8K    & \bf 16K  & \bf 32K  & \bf 64K   & \bf 128K  \\ \midrule
& {Qwen2.5-7B-Instruct}  & 85.4 & 96.7 & 95.1 &93.7 & 89.4&82.3&55.1\\
& {Qwen2.5-14B-Instruct} & 91.4 &97.7 &96.8 &95.9 &93.4 & 86.7 & 78.1 \\
& {Qwen2.5-32B-Instruct} & 92.9 & 96.9 & 97.1 & 95.5 & 95.5 & 90.3 & 82.0  \\
& {Qwen2.5-72B-Instruct} & \bf95.1 & \bf97.7 & \bf 97.2 & \bf 97.7 & \bf 96.5 & 93.0 & 88.4 \\
\midrule
\multirow{6}{*}{\tabincell{c}{\textit{Non-thinking}\\\textit{Mode}}} & \textbf{Qwen3-4B} & 85.2 & 95.1 & 93.6 & 91.0 & 87.8 & 77.8 & 66.0 \\
& \textbf{Qwen3-8B} & 89.1 & 96.3 & 96.0 & 91.8 & 91.2 & 82.1 & 77.4 \\
& \textbf{Qwen3-14B} & 94.6 & 98.0 & 97.8 & 96.4 & 96.1 & 94.0 & 85.1 \\
& \textbf{Qwen3-32B} & 93.7 & 98.4 & 96.0 & 96.2 & 94.4 & 91.8 & 85.6 \\
& \textbf{Qwen3-30B-A3B} & 91.6 & 96.5 & 97.0 & 95.3 & 92.4 & 89.1 & 79.2 \\
& \textbf{Qwen3-235B-A22B} & 95.0 & \bf97.7 & \bf97.2 & 96.4 & 95.1 & \bf93.3 & \bf90.6 \\
\midrule
\multirow{6}{*}{\tabincell{c}{\textit{Thinking}\\\textit{Mode}}} & \textbf{Qwen3-4B} & 83.5 & 92.7 & 88.7 & 86.5 & 83.2 & 83.0 & 67.2 \\
& \textbf{Qwen3-8B} & 84.4 & 94.7 & 94.4 & 86.1 & 80.8 & 78.3 & 72.0 \\
& \textbf{Qwen3-14B} & 90.1 & 95.4 & 93.6 & 89.8 & 91.9 & 90.6 & 79.0 \\
& \textbf{Qwen3-32B} & 91.0 & 94.7 & 93.7 & 91.6 & 92.5 & 90.0 & 83.5 \\
& \textbf{Qwen3-30B-A3B} & 86.6 & 94.1 & 92.7 & 89.0 & 86.6 & 82.1 & 75.0 \\
& \textbf{Qwen3-235B-A22B} & 92.2 & 95.1 & 94.8 & 93.0 & 92.3 & 92.0 & 86.0 \\
\bottomrule
\end{tabular}
\end{table}

For evaluating long-context processing capabilities, we report the results on the RULER benchmark~\citep{hsieh2024ruler} in Table~\ref{tab:ruler}. To enable length extrapolation, we utilize YARN~\citep{yarn} with a \texttt{scaling\_factor=4}. In thinking mode, we set the thinking budget to 8192 tokens to mitigate overly verbose reasoning on the extremely long inputs.

The results show that:
\begin{enumerate}
    \item In non-thinking mode, Qwen3 outperforms Qwen2.5 models of a similar size in long-context processing tasks.
    \item In thinking mode, the model's performance slightly degrades. We hypothesize that the thinking content does not provide significant benefits for these retrieval tasks,  which do not rely on reasoning and may instead interfere with the retrieval process. We are committed to enhancing the long-context capability in the thinking mode in future versions.
\end{enumerate}

\subsubsection{Multilingual Ability}
Table \ref{tab:scores_es}-\ref{tab:scores_th} presents the detailed benchmark scores across various languages, including Spanish, French, Portuguese, Italian, Arabic, Japanese, Korean, Indonesian, Russian, Vietnamese, German, and Thai. The results of these tables demonstrate that the Qwen3 series models achieve competitive performance across all evaluated benchmarks, showcasing their strong multilingual capabilities.

To evaluate the performance of Qwen3 across a broader range of languages, we utilize Belebele~\citep{belebele}, a benchmark for natural language understanding. We conduct evaluations on 80 supported languages from the benchmark, excluding 42 unoptimized languages, as shown in Table \ref{tab:Belebele_language_codes} (organized by language family). The performance comparison between Qwen3 and other baseline models on the Belebele benchmark is presented in Table \ref{tab:belebele}. The results show that Qwen3 achieves comparable performance to similarly-sized Gemma models while outperforming Qwen2.5 significantly.

\begin{table}[tbp]
\centering
\caption{\textbf{Benchmark scores for language: Spanish (es)}. The highest and second-best scores are shown in \textbf{bold} and \underline{underlined}, respectively.}
\label{tab:scores_es}

\setlength\tabcolsep{2.5pt}
\adjustbox{center=\textwidth}{
\small
\begin{tabular}{clccccccc}
\toprule
 & Model & Multi-IF & MLogiQA & INCLUDE & MMMLU & MT-AIME24 & PolyMath & Average \\
\midrule
\multirow{10}{*}{\tabincell{c}{\textit{Thinking}\\\textit{Mode}}} & Gemini2.5-Pro & \textbf{80.1} & 70.0 & \textbf{96.4} & \textbf{88.7} & \textbf{90.0} & \underline{54.4} & \textbf{79.9} \\
 & QwQ-32B & 70.0 & \underline{75.0} & 81.8 & 84.5 & 76.7 & 52.2 & 73.4 \\
 & \textbf{Qwen3-235B-A22B} & 74.2 & \textbf{76.2} & 89.1 & \underline{86.7} & \underline{86.7} & \textbf{57.3} & \underline{78.4} \\
 & \textbf{Qwen3-32B} & 74.7 & 68.8 & \underline{90.9} & 82.8 & 76.7 & 51.8 & 74.3 \\
 & \textbf{Qwen3-30B-A3B} & 74.9 & 71.2 & 80.0 & 81.9 & 76.7 & 48.5 & 72.2 \\
 & \textbf{Qwen3-14B} & \underline{76.2} & 67.5 & 83.6 & 81.1 & 73.3 & 50.3 & 72.0 \\
 & \textbf{Qwen3-8B} & 74.1 & 70.0 & 78.2 & 79.2 & 70.0 & 43.7 & 69.2 \\
\bottomrule
\end{tabular}
}
\end{table}

\begin{table}[tbp]
\centering
\caption{\textbf{Benchmark scores for language: French (fr)}. The highest and second-best scores are shown in \textbf{bold} and \underline{underlined}, respectively.}
\label{tab:scores_fr}
\setlength\tabcolsep{2.5pt}
\adjustbox{center=\textwidth}{
\small
\begin{tabular}{clccccccc}
\toprule
 & Model & Multi-IF & MLogiQA & INCLUDE & MMMLU & MT-AIME24 & PolyMath & Average \\
\midrule
\multirow{10}{*}{\tabincell{c}{\textit{Thinking}\\\textit{Mode}}} & Gemini2.5-Pro & \textbf{80.5} & 73.8 & \textbf{85.7} & \textbf{88.3} & 80.0 & \underline{52.8} & \underline{76.8} \\
 & QwQ-32B & 72.4 & \underline{78.8} & 76.2 & 84.0 & 80.0 & 49.4 & 73.5 \\
 & \textbf{Qwen3-235B-A22B} & 77.3 & \underline{78.8} & \textbf{85.7} & \underline{86.6} & \textbf{86.7} & \textbf{57.4} & \textbf{78.8} \\
 & \textbf{Qwen3-32B} & 76.7 & \textbf{81.2} & 76.2 & 82.1 & \underline{83.3} & 47.1 & 74.4 \\
 & \textbf{Qwen3-30B-A3B} & 75.2 & 67.5 & \underline{83.3} & 81.0 & 76.7 & 46.9 & 71.8 \\
 & \textbf{Qwen3-14B} & \underline{77.6} & 71.2 & 73.8 & 80.4 & 73.3 & 44.2 & 70.1 \\
 & \textbf{Qwen3-8B} & 73.8 & 66.2 & \textbf{85.7} & 77.9 & 70.0 & 45.3 & 69.8 \\
\bottomrule
\end{tabular}
}
\end{table}

\begin{table}[tbp]
\centering
\caption{\textbf{Benchmark scores for language: Portuguese (pt)}. The highest and second-best scores are shown in \textbf{bold} and \underline{underlined}, respectively.}
\label{tab:scores_pt}
\setlength\tabcolsep{2.5pt}
\adjustbox{center=\textwidth}{
\small
\begin{tabular}{clccccccc}
\toprule
 & Model & Multi-IF & MLogiQA & INCLUDE & MMMLU & MT-AIME24 & PolyMath & Average \\
\midrule
\multirow{10}{*}{\tabincell{c}{\textit{Thinking}\\\textit{Mode}}} & Gemini2.5-Pro & \textbf{80.5} & 73.8 & \textbf{83.9} & \textbf{88.9} & 73.3 & 52.2 & \underline{75.4} \\
 & QwQ-32B & 70.5 & 70.0 & \underline{80.4} & 84.0 & 80.0 & 48.7 & 72.3 \\
 & \textbf{Qwen3-235B-A22B} & 73.6 & \textbf{78.8} & 78.6 & \underline{86.2} & \textbf{86.7} & \textbf{58.3} & \textbf{77.0} \\
 & \textbf{Qwen3-32B} & 74.1 & \underline{76.2} & 76.8 & 82.6 & 80.0 & \underline{52.4} & 73.7 \\
 & \textbf{Qwen3-30B-A3B} & 76.1 & 71.2 & 71.4 & 81.0 & 76.7 & 49.3 & 71.0 \\
 & \textbf{Qwen3-14B} & \underline{77.3} & 68.8 & 75.0 & 81.6 & \underline{83.3} & 46.7 & 72.1 \\
 & \textbf{Qwen3-8B} & 73.9 & 67.5 & 75.0 & 78.6 & 56.7 & 44.8 & 66.1 \\
\bottomrule
\end{tabular}
}
\end{table}

\begin{table}[tbp]
\centering
\caption{\textbf{Benchmark scores for language: Italian (it)}. The highest and second-best scores are shown in \textbf{bold} and \underline{underlined}, respectively.}
\label{tab:scores_it}
\setlength\tabcolsep{2.5pt}
\adjustbox{center=\textwidth}{
\small
\begin{tabular}{clcccccc}
\toprule
 & Model & Multi-IF & INCLUDE & MMMLU & MT-AIME24 & PolyMath & Average \\
\midrule
\multirow{10}{*}{\tabincell{c}{\textit{Thinking}\\\textit{Mode}}} & Gemini2.5-Pro & \textbf{80.9} & \textbf{100.0} & \textbf{87.2} & \textbf{90.0} & \underline{54.1} & \textbf{82.4} \\
 & QwQ-32B & 71.2 & \underline{96.4} & 84.9 & 76.7 & 49.3 & 75.7 \\
 & \textbf{Qwen3-235B-A22B} & 73.7 & \underline{96.4} & \underline{85.7} & \underline{80.0} & \textbf{57.4} & \underline{78.6} \\
 & \textbf{Qwen3-32B} & 76.6 & 90.9 & 81.6 & \underline{80.0} & 49.7 & 75.8 \\
 & \textbf{Qwen3-30B-A3B} & 75.9 & 94.5 & 81.9 & \underline{80.0} & 48.1 & 76.1 \\
 & \textbf{Qwen3-14B} & \underline{79.0} & 94.5 & 80.2 & 70.0 & 47.0 & 74.1 \\
 & \textbf{Qwen3-8B} & 74.6 & 89.1 & 77.5 & 76.7 & 46.1 & 72.8 \\
 & \textbf{Qwen3-4B} & 69.8 & 83.6 & 74.4 & 76.7 & 44.5 & 69.8 \\
 & \textbf{Qwen3-1.7B} & 54.6 & 74.5 & 64.2 & 53.3 & 29.6 & 55.2 \\
 & \textbf{Qwen3-0.6B} & 37.8 & 45.5 & 45.9 & 6.7 & 13.3 & 29.8 \\
\midrule
\multirow{11}{*}{\tabincell{c}{\textit{Non-thinking}\\\textit{Mode}}} & GPT-4o-2024-1120 & 67.6 & \textbf{98.2} & \underline{80.7} & 13.3 & 15.2 & 55.0 \\
 & Gemma-3-27b-IT & \underline{74.6} & 90.9 & 78.4 & 23.3 & 20.5 & 57.5 \\
 & Qwen2.5-72B-Instruct & 67.2 & \underline{94.5} & \underline{80.7} & 16.7 & 16.7 & 55.2 \\
 & \textbf{Qwen3-235B-A22B} & 72.9 & 92.7 & \textbf{82.6} & \textbf{33.3} & \textbf{28.6} & \textbf{62.0} \\
 & \textbf{Qwen3-32B} & 71.4 & 92.7 & 79.5 & \underline{30.0} & 23.0 & 59.3 \\
 & \textbf{Qwen3-30B-A3B} & 73.9 & 87.3 & 77.7 & \textbf{33.3} & 24.8 & \underline{59.4} \\
 & \textbf{Qwen3-14B} & \textbf{75.8} & 89.1 & 75.7 & 26.7 & \underline{27.6} & 59.0 \\
 & \textbf{Qwen3-8B} & 72.1 & 85.5 & 72.9 & 13.3 & 23.8 & 53.5 \\
 & \textbf{Qwen3-4B} & 63.0 & 78.2 & 67.8 & 23.3 & 19.3 & 50.3 \\
 & \textbf{Qwen3-1.7B} & 46.1 & 70.9 & 53.4 & 6.7 & 11.9 & 37.8 \\
 & \textbf{Qwen3-0.6B} & 35.1 & 43.6 & 39.0 & 0.0 & 4.5 & 24.4 \\
\bottomrule
\end{tabular}
}
\end{table}

\begin{table}[tbp]
\centering
\caption{\textbf{Benchmark scores for language: Arabic (ar)}. The highest and second-best scores are shown in \textbf{bold} and \underline{underlined}, respectively.}
\label{tab:scores_ar}
\setlength\tabcolsep{2.5pt}
\adjustbox{center=\textwidth}{
\small
\begin{tabular}{clcccccc}
\toprule
 & Model & MLogiQA & INCLUDE & MMMLU & MT-AIME24 & PolyMath & Average \\
\midrule
\multirow{10}{*}{\tabincell{c}{\textit{Thinking}\\\textit{Mode}}} & Gemini2.5-Pro & \underline{75.0} & \textbf{89.3} & \textbf{87.8} & 76.7 & \underline{52.6} & \textbf{76.3} \\
 & QwQ-32B & \underline{75.0} & 67.9 & 81.8 & 80.0 & 41.3 & 69.2 \\
 & \textbf{Qwen3-235B-A22B} & \textbf{80.0} & 71.4 & \underline{83.6} & 76.7 & \textbf{53.7} & \underline{73.1} \\
 & \textbf{Qwen3-32B} & 66.2 & \underline{73.2} & 80.1 & \textbf{86.7} & 47.0 & 70.6 \\
 & \textbf{Qwen3-30B-A3B} & 66.2 & 66.1 & 77.2 & \underline{83.3} & 47.3 & 68.0 \\
 & \textbf{Qwen3-14B} & 71.2 & 67.9 & 77.4 & \underline{83.3} & 46.6 & 69.3 \\
 & \textbf{Qwen3-8B} & 65.0 & 67.9 & 74.4 & 76.7 & 44.9 & 65.8 \\
 & \textbf{Qwen3-4B} & 62.5 & 55.4 & 67.7 & 66.7 & 41.2 & 58.7 \\
 & \textbf{Qwen3-1.7B} & 55.0 & 44.6 & 53.2 & 36.7 & 25.8 & 43.1 \\
 & \textbf{Qwen3-0.6B} & 40.0 & 41.1 & 38.9 & 10.0 & 11.7 & 28.3 \\
\midrule
\multirow{11}{*}{\tabincell{c}{\textit{Non-thinking}\\\textit{Mode}}} & GPT-4o-2024-1120 & 51.2 & \textbf{78.6} & \textbf{80.9} & 13.3 & 12.9 & 47.4 \\
 & Gemma-3-27b-IT & \underline{56.2} & 62.5 & 74.4 & 26.7 & 22.8 & 48.5 \\
 & Qwen2.5-72B-Instruct & \underline{56.2} & 66.1 & 77.2 & 6.7 & 14.7 & 44.2 \\
 & \textbf{Qwen3-235B-A22B} & \textbf{66.2} & 67.9 & \underline{79.5} & \textbf{40.0} & \textbf{28.2} & \textbf{56.4} \\
 & \textbf{Qwen3-32B} & 55.0 & \underline{69.6} & 75.7 & 23.3 & \underline{25.4} & \underline{49.8} \\
 & \textbf{Qwen3-30B-A3B} & 48.8 & 64.3 & 71.6 & \underline{30.0} & 22.6 & 47.5 \\
 & \textbf{Qwen3-14B} & 52.5 & 60.7 & 69.5 & 23.3 & 23.5 & 45.9 \\
 & \textbf{Qwen3-8B} & 45.0 & 58.9 & 64.6 & 13.3 & 16.4 & 39.6 \\
 & \textbf{Qwen3-4B} & 52.5 & 42.9 & 56.7 & 13.3 & 15.3 & 36.1 \\
 & \textbf{Qwen3-1.7B} & 31.2 & 37.5 & 43.6 & 3.3 & 9.4 & 25.0 \\
 & \textbf{Qwen3-0.6B} & 40.0 & 39.3 & 35.4 & 0.0 & 3.8 & 23.7 \\
\bottomrule
\end{tabular}
}
\end{table}

\begin{table}[tbp]
\centering
\caption{\textbf{Benchmark scores for language: Japanese (ja)}. The highest and second-best scores are shown in \textbf{bold} and \underline{underlined}, respectively.}
\label{tab:scores_ja}
\setlength\tabcolsep{2.5pt}
\adjustbox{center=\textwidth}{
\small
\begin{tabular}{clcccccc}
\toprule
 & Model & MLogiQA & INCLUDE & MMMLU & MT-AIME24 & PolyMath & Average \\
\midrule
\multirow{10}{*}{\tabincell{c}{\textit{Thinking}\\\textit{Mode}}} & Gemini2.5-Pro & 72.5 & 74.5 & \underline{83.8} & \textbf{83.3} & \textbf{55.4} & \underline{73.9} \\
 & QwQ-32B & \underline{73.8} & 86.3 & 82.3 & 53.3 & 39.9 & 67.1 \\
 & \textbf{Qwen3-235B-A22B} & \textbf{75.0} & \textbf{94.1} & \textbf{84.8} & 73.3 & \underline{52.7} & \textbf{76.0} \\
 & \textbf{Qwen3-32B} & 70.0 & \underline{90.2} & 80.2 & \underline{76.7} & 47.7 & 73.0 \\
 & \textbf{Qwen3-30B-A3B} & 66.2 & 88.2 & 79.9 & 73.3 & 47.4 & 71.0 \\
 & \textbf{Qwen3-14B} & 68.8 & 88.2 & 79.4 & 66.7 & 45.7 & 69.8 \\
 & \textbf{Qwen3-8B} & 71.2 & 86.3 & 74.9 & 73.3 & 44.7 & 70.1 \\
 & \textbf{Qwen3-4B} & 63.7 & 80.4 & 72.5 & 53.3 & 40.7 & 62.1 \\
 & \textbf{Qwen3-1.7B} & 53.8 & 74.5 & 61.8 & 36.7 & 28.5 & 51.1 \\
 & \textbf{Qwen3-0.6B} & 47.5 & 47.1 & 45.1 & 13.3 & 14.5 & 33.5 \\
\midrule
\multirow{11}{*}{\tabincell{c}{\textit{Non-thinking}\\\textit{Mode}}} & GPT-4o-2024-1120 & 60.0 & \underline{92.2} & \textbf{81.9} & 10.0 & 12.5 & 51.3 \\
 & Gemma-3-27b-IT & \underline{66.2} & 86.3 & 76.5 & 20.0 & 17.3 & 53.3 \\
 & Qwen2.5-72B-Instruct & 55.0 & \textbf{94.1} & 77.7 & 16.7 & 17.7 & 52.2 \\
 & \textbf{Qwen3-235B-A22B} & \textbf{67.5} & \underline{92.2} & \underline{80.9} & 26.7 & \textbf{26.9} & \textbf{58.8} \\
 & \textbf{Qwen3-32B} & 58.8 & \underline{92.2} & 78.0 & 20.0 & 20.5 & \underline{53.9} \\
 & \textbf{Qwen3-30B-A3B} & 51.2 & 82.4 & 74.9 & \underline{30.0} & \underline{20.6} & 51.8 \\
 & \textbf{Qwen3-14B} & 55.0 & 84.3 & 73.8 & \textbf{33.3} & 19.8 & 53.2 \\
 & \textbf{Qwen3-8B} & 47.5 & 82.4 & 69.9 & 20.0 & 18.5 & 47.7 \\
 & \textbf{Qwen3-4B} & 46.2 & 76.5 & 64.8 & 13.3 & 15.1 & 43.2 \\
 & \textbf{Qwen3-1.7B} & 40.0 & 68.6 & 46.3 & 3.3 & 11.6 & 34.0 \\
 & \textbf{Qwen3-0.6B} & 37.5 & 37.3 & 37.9 & 3.3 & 3.7 & 23.9 \\
\bottomrule
\end{tabular}
}
\end{table}

\begin{table}[tbp]
\centering
\caption{\textbf{Benchmark scores for language: Korean (ko)}. The highest and second-best scores are shown in \textbf{bold} and \underline{underlined}, respectively.}
\label{tab:scores_ko}
\setlength\tabcolsep{2.5pt}
\adjustbox{center=\textwidth}{
\small
\begin{tabular}{clcccccc}
\toprule
 & Model & MLogiQA & INCLUDE & MMMLU & MT-AIME24 & PolyMath & Average \\
\midrule
\multirow{10}{*}{\tabincell{c}{\textit{Thinking}\\\textit{Mode}}} & Gemini2.5-Pro & \underline{75.0} & \textbf{88.0} & \textbf{85.9} & \underline{76.7} & \underline{50.0} & \textbf{75.1} \\
 & QwQ-32B & \textbf{76.2} & 72.0 & 81.8 & 60.0 & 40.0 & 66.0 \\
 & \textbf{Qwen3-235B-A22B} & 71.2 & \underline{80.0} & \underline{84.7} & \textbf{80.0} & \textbf{55.7} & \underline{74.3} \\
 & \textbf{Qwen3-32B} & 71.2 & 74.0 & 79.2 & \textbf{80.0} & 48.5 & 70.6 \\
 & \textbf{Qwen3-30B-A3B} & 68.8 & 72.0 & 78.6 & \underline{76.7} & 46.6 & 68.5 \\
 & \textbf{Qwen3-14B} & 67.5 & 74.0 & 79.6 & \underline{76.7} & 46.0 & 68.8 \\
 & \textbf{Qwen3-8B} & 60.0 & \underline{80.0} & 74.7 & \underline{76.7} & 42.3 & 66.7 \\
\bottomrule
\end{tabular}
}
\end{table}

\begin{table}[tbp]
\centering
\caption{\textbf{Benchmark scores for language: Indonesian (id)}. The highest and second-best scores are shown in \textbf{bold} and \underline{underlined}, respectively.}
\label{tab:scores_id}
\setlength\tabcolsep{2.5pt}
\adjustbox{center=\textwidth}{
\small
\begin{tabular}{clccccc}
\toprule
 & Model & INCLUDE & MMMLU & MT-AIME24 & PolyMath & Average \\
\midrule
\multirow{10}{*}{\tabincell{c}{\textit{Thinking}\\\textit{Mode}}} & Gemini2.5-Pro & \underline{80.0} & \underline{86.3} & \textbf{83.3} & \underline{51.3} & \textbf{75.2} \\
 & QwQ-32B & 76.4 & 83.7 & 73.3 & 47.3 & 70.2 \\
 & \textbf{Qwen3-235B-A22B} & \underline{80.0} & \textbf{87.2} & \underline{80.0} & \textbf{53.5} & \textbf{75.2} \\
 & \textbf{Qwen3-32B} & \underline{80.0} & 82.0 & 76.7 & 45.6 & 71.1 \\
 & \textbf{Qwen3-30B-A3B} & \textbf{81.8} & 80.4 & \underline{80.0} & 44.9 & \underline{71.8} \\
 & \textbf{Qwen3-14B} & 78.2 & 79.6 & 70.0 & 45.3 & 68.3 \\
 & \textbf{Qwen3-8B} & 72.7 & 77.7 & 70.0 & 43.8 & 66.0 \\
 & \textbf{Qwen3-4B} & 70.9 & 72.3 & 66.7 & 41.2 & 62.8 \\
 & \textbf{Qwen3-1.7B} & 63.6 & 61.2 & 36.7 & 26.8 & 47.1 \\
 & \textbf{Qwen3-0.6B} & 36.4 & 46.6 & 10.0 & 12.6 & 26.4 \\
\midrule
\multirow{11}{*}{\tabincell{c}{\textit{Non-thinking}\\\textit{Mode}}} & GPT-4o-2024-1120 & \underline{80.0} & \underline{81.1} & 10.0 & 14.7 & 46.4 \\
 & Gemma-3-27b-IT & 76.4 & 75.9 & 13.3 & 22.6 & 47.0 \\
 & Qwen2.5-72B-Instruct & 74.5 & 78.8 & 10.0 & 16.6 & 45.0 \\
 & \textbf{Qwen3-235B-A22B} & \textbf{81.8} & \textbf{81.9} & \textbf{33.3} & \textbf{27.5} & \textbf{56.1} \\
 & \textbf{Qwen3-32B} & \textbf{81.8} & 77.2 & 23.3 & 24.3 & \underline{51.6} \\
 & \textbf{Qwen3-30B-A3B} & 70.9 & 76.4 & \underline{30.0} & \underline{25.9} & 50.8 \\
 & \textbf{Qwen3-14B} & 70.9 & 74.1 & 26.7 & 24.6 & 49.1 \\
 & \textbf{Qwen3-8B} & 78.2 & 69.6 & 20.0 & 21.6 & 47.4 \\
 & \textbf{Qwen3-4B} & 67.3 & 66.5 & 13.3 & 19.0 & 41.5 \\
 & \textbf{Qwen3-1.7B} & 52.7 & 49.0 & 3.3 & 10.8 & 29.0 \\
 & \textbf{Qwen3-0.6B} & 52.7 & 40.0 & 3.3 & 5.1 & 25.3 \\
\bottomrule
\end{tabular}
}
\end{table}

\begin{table}[tbp]
\centering
\caption{\textbf{Benchmark scores for language: Russian (ru)}. The highest and second-best scores are shown in \textbf{bold} and \underline{underlined}, respectively.}
\label{tab:scores_ru}
\setlength\tabcolsep{2.5pt}
\adjustbox{center=\textwidth}{
\small
\begin{tabular}{clccccc}
\toprule
 & Model & Multi-IF & INCLUDE & MT-AIME24 & PolyMath & Average \\
\midrule
\multirow{10}{*}{\tabincell{c}{\textit{Thinking}\\\textit{Mode}}} & Gemini2.5-Pro & \textbf{68.1} & \textbf{80.4} & 70.0 & \underline{52.3} & \underline{67.7} \\
 & QwQ-32B & 61.2 & 73.2 & \underline{76.7} & 43.6 & 63.7 \\
 & \textbf{Qwen3-235B-A22B} & 62.2 & \textbf{80.4} & \textbf{80.0} & \textbf{53.1} & \textbf{68.9} \\
 & \textbf{Qwen3-32B} & 62.5 & 73.2 & 63.3 & 46.5 & 61.4 \\
 & \textbf{Qwen3-30B-A3B} & 60.7 & \underline{76.8} & 73.3 & 45.4 & 64.0 \\
 & \textbf{Qwen3-14B} & \underline{63.6} & \textbf{80.4} & 66.7 & 46.4 & 64.3 \\
 & \textbf{Qwen3-8B} & 62.9 & 69.6 & 63.3 & 37.7 & 58.4 \\
 & \textbf{Qwen3-4B} & 52.8 & 69.6 & 56.7 & 36.6 & 53.9 \\
 & \textbf{Qwen3-1.7B} & 37.8 & 46.4 & 20.0 & 22.8 & 31.8 \\
 & \textbf{Qwen3-0.6B} & 26.4 & 46.4 & 3.3 & 7.0 & 20.8 \\
\midrule
\multirow{11}{*}{\tabincell{c}{\textit{Non-thinking}\\\textit{Mode}}} & GPT-4o-2024-1120 & 52.0 & \textbf{80.4} & 20.0 & 13.7 & 41.5 \\
 & Gemma-3-27b-IT & 57.3 & 71.4 & 23.3 & 21.6 & 43.4 \\
 & Qwen2.5-72B-Instruct & 54.1 & 67.9 & 20.0 & 13.3 & 38.8 \\
 & \textbf{Qwen3-235B-A22B} & 56.7 & \underline{75.0} & \textbf{40.0} & \textbf{26.1} & \textbf{49.4} \\
 & \textbf{Qwen3-32B} & 58.6 & 71.4 & \underline{30.0} & 23.3 & \underline{45.8} \\
 & \textbf{Qwen3-30B-A3B} & 58.0 & 73.2 & \underline{30.0} & 21.1 & 45.6 \\
 & \textbf{Qwen3-14B} & \textbf{60.3} & 71.4 & 26.7 & \underline{24.2} & 45.6 \\
 & \textbf{Qwen3-8B} & \underline{59.3} & 58.9 & 20.0 & 22.8 & 40.2 \\
 & \textbf{Qwen3-4B} & 46.1 & 58.9 & 13.3 & 17.8 & 34.0 \\
 & \textbf{Qwen3-1.7B} & 34.8 & 41.1 & 3.3 & 13.2 & 23.1 \\
 & \textbf{Qwen3-0.6B} & 25.5 & 46.4 & 0.0 & 5.8 & 19.4 \\
\bottomrule
\end{tabular}
}
\end{table}

\begin{table}[tbp]
\centering
\caption{\textbf{Benchmark scores for language: Vietnamese (vi)}. The highest and second-best scores are shown in \textbf{bold} and \underline{underlined}, respectively.}
\label{tab:scores_vi}
\setlength\tabcolsep{2.5pt}
\adjustbox{center=\textwidth}{
\small
\begin{tabular}{clccccc}
\toprule
 & Model & MLogiQA & INCLUDE & MT-AIME24 & PolyMath & Average \\
\midrule
\multirow{10}{*}{\tabincell{c}{\textit{Thinking}\\\textit{Mode}}} & Gemini2.5-Pro & \underline{72.5} & \textbf{89.1} & 70.0 & \underline{52.1} & \underline{70.9} \\
 & QwQ-32B & 71.2 & 69.1 & 70.0 & 49.2 & 64.9 \\
 & \textbf{Qwen3-235B-A22B} & \textbf{75.0} & \underline{87.3} & \textbf{83.3} & \textbf{55.1} & \textbf{75.2} \\
 & \textbf{Qwen3-32B} & 67.5 & 81.8 & \textbf{83.3} & 44.0 & 69.2 \\
 & \textbf{Qwen3-30B-A3B} & 68.8 & 78.2 & \underline{76.7} & 46.1 & 67.4 \\
 & \textbf{Qwen3-14B} & \underline{72.5} & 72.7 & 73.3 & 45.8 & 66.1 \\
 & \textbf{Qwen3-8B} & 65.0 & 72.7 & 73.3 & 42.9 & 63.5 \\
 & \textbf{Qwen3-4B} & 68.8 & 63.6 & 60.0 & 42.2 & 58.6 \\
 & \textbf{Qwen3-1.7B} & 52.5 & 61.8 & 30.0 & 26.9 & 42.8 \\
 & \textbf{Qwen3-0.6B} & 33.8 & 38.2 & 6.7 & 9.8 & 22.1 \\
\midrule
\multirow{11}{*}{\tabincell{c}{\textit{Non-thinking}\\\textit{Mode}}} & GPT-4o-2024-1120 & 57.5 & \underline{81.8} & 10.0 & 13.0 & 40.6 \\
 & Gemma-3-27b-IT & 52.5 & 74.5 & \underline{33.3} & 20.6 & 45.2 \\
 & Qwen2.5-72B-Instruct & 61.3 & 72.7 & 26.7 & 18.6 & 44.8 \\
 & \textbf{Qwen3-235B-A22B} & \textbf{70.0} & \textbf{83.6} & \textbf{36.7} & \textbf{27.1} & \textbf{54.4} \\
 & \textbf{Qwen3-32B} & 60.0 & \underline{81.8} & 23.3 & 21.8 & \underline{46.7} \\
 & \textbf{Qwen3-30B-A3B} & 52.5 & \underline{81.8} & 20.0 & \underline{24.7} & 44.8 \\
 & \textbf{Qwen3-14B} & \underline{63.7} & 67.3 & 20.0 & 21.6 & 43.2 \\
 & \textbf{Qwen3-8B} & 48.8 & 65.5 & 20.0 & 19.1 & 38.4 \\
 & \textbf{Qwen3-4B} & 48.8 & 65.5 & 20.0 & 19.0 & 38.3 \\
 & \textbf{Qwen3-1.7B} & 36.2 & 60.0 & 3.3 & 10.9 & 27.6 \\
 & \textbf{Qwen3-0.6B} & 30.0 & 36.4 & 3.3 & 3.9 & 18.4 \\
\bottomrule
\end{tabular}
}
\end{table}

\begin{table}[tbp]
\centering
\caption{\textbf{Benchmark scores for language: German (de)}. The highest and second-best scores are shown in \textbf{bold} and \underline{underlined}, respectively.}
\label{tab:scores_de}
\setlength\tabcolsep{2.5pt}
\adjustbox{center=\textwidth}{
\small
\begin{tabular}{clccccc}
\toprule
 & Model & INCLUDE & MMMLU & MT-AIME24 & PolyMath & Average \\
\midrule
\multirow{10}{*}{\tabincell{c}{\textit{Thinking}\\\textit{Mode}}} & Gemini2.5-Pro & 50.0 & \underline{85.6} & \textbf{86.7} & \underline{53.8} & 69.0 \\
 & QwQ-32B & 57.1 & 83.8 & 76.7 & 51.0 & 67.2 \\
 & \textbf{Qwen3-235B-A22B} & \textbf{71.4} & \textbf{86.0} & \underline{83.3} & \textbf{55.4} & \textbf{74.0} \\
 & \textbf{Qwen3-32B} & \underline{64.3} & 81.9 & \textbf{86.7} & 48.1 & \underline{70.2} \\
 & \textbf{Qwen3-30B-A3B} & \underline{64.3} & 81.9 & 80.0 & 46.6 & 68.2 \\
 & \textbf{Qwen3-14B} & 57.1 & 80.9 & 70.0 & 48.1 & 64.0 \\
 & \textbf{Qwen3-8B} & \underline{64.3} & 78.1 & 66.7 & 43.6 & 63.2 \\
 & \textbf{Qwen3-4B} & 57.1 & 74.0 & 73.3 & 43.1 & 61.9 \\
 & \textbf{Qwen3-1.7B} & \underline{64.3} & 63.4 & 36.7 & 26.8 & 47.8 \\
 & \textbf{Qwen3-0.6B} & 57.1 & 47.6 & 10.0 & 13.7 & 32.1 \\
\midrule
\multirow{11}{*}{\tabincell{c}{\textit{Non-thinking}\\\textit{Mode}}} & GPT-4o-2024-1120 & 57.1 & \underline{80.4} & 10.0 & 13.5 & 40.2 \\
 & Gemma-3-27b-IT & 57.1 & 76.1 & 26.7 & 20.2 & 45.0 \\
 & Qwen2.5-72B-Instruct & \underline{64.3} & 79.9 & 16.7 & 19.3 & 45.0 \\
 & \textbf{Qwen3-235B-A22B} & \textbf{71.4} & \textbf{81.7} & \textbf{40.0} & \textbf{25.9} & \textbf{54.8} \\
 & \textbf{Qwen3-32B} & 57.1 & 77.2 & \underline{30.0} & 21.9 & 46.6 \\
 & \textbf{Qwen3-30B-A3B} & 57.1 & 77.7 & 23.3 & \underline{25.2} & 45.8 \\
 & \textbf{Qwen3-14B} & 57.1 & 76.0 & \underline{30.0} & 24.5 & \underline{46.9} \\
 & \textbf{Qwen3-8B} & \underline{64.3} & 70.8 & 20.0 & 19.9 & 43.8 \\
 & \textbf{Qwen3-4B} & \underline{64.3} & 66.0 & 26.7 & 16.4 & 43.4 \\
 & \textbf{Qwen3-1.7B} & 42.9 & 53.2 & 10.0 & 10.6 & 29.2 \\
 & \textbf{Qwen3-0.6B} & 42.9 & 37.8 & 3.3 & 5.7 & 22.4 \\
\bottomrule
\end{tabular}
}
\end{table}

\begin{table}[tbp]
\centering
\caption{\textbf{Benchmark scores for language: Thai (th)}. The highest and second-best scores are shown in \textbf{bold} and \underline{underlined}, respectively.}
\label{tab:scores_th}
\setlength\tabcolsep{2.5pt}
\adjustbox{center=\textwidth}{
\small
\begin{tabular}{clcccc}
\toprule
 & Model & MLogiQA & MT-AIME24 & PolyMath & Average \\
\midrule
\multirow{10}{*}{\tabincell{c}{\textit{Thinking}\\\textit{Mode}}} & Gemini2.5-Pro & \underline{73.8} & \underline{80.0} & \underline{50.7} & \underline{68.2} \\
 & QwQ-32B & \textbf{75.0} & 60.0 & 41.3 & 58.8 \\
 & \textbf{Qwen3-235B-A22B} & \underline{73.8} & \textbf{86.7} & \textbf{53.6} & \textbf{71.4} \\
 & \textbf{Qwen3-32B} & \underline{73.8} & 76.7 & 46.9 & 65.8 \\
 & \textbf{Qwen3-30B-A3B} & 63.7 & \underline{80.0} & 45.2 & 63.0 \\
 & \textbf{Qwen3-14B} & 65.0 & 76.7 & 44.4 & 62.0 \\
 & \textbf{Qwen3-8B} & 68.8 & 70.0 & 41.3 & 60.0 \\
 & \textbf{Qwen3-4B} & 60.0 & 60.0 & 39.4 & 53.1 \\
 & \textbf{Qwen3-1.7B} & 48.8 & 33.3 & 23.7 & 35.3 \\
 & \textbf{Qwen3-0.6B} & 33.8 & 13.3 & 11.4 & 19.5 \\
\midrule
\multirow{11}{*}{\tabincell{c}{\textit{Non-thinking}\\\textit{Mode}}} & GPT-4o-2024-1120 & 52.5 & 10.0 & 11.9 & 24.8 \\
 & Gemma-3-27b-IT & 50.0 & 16.7 & 19.0 & 28.6 \\
 & Qwen2.5-72B-Instruct & \underline{58.8} & 6.7 & 17.4 & 27.6 \\
 & \textbf{Qwen3-235B-A22B} & \textbf{61.3} & \underline{23.3} & \textbf{27.6} & \textbf{37.4} \\
 & \textbf{Qwen3-32B} & \textbf{61.3} & 13.3 & 22.2 & 32.3 \\
 & \textbf{Qwen3-30B-A3B} & 50.0 & \textbf{30.0} & \underline{22.3} & \underline{34.1} \\
 & \textbf{Qwen3-14B} & 47.5 & \underline{23.3} & 22.1 & 31.0 \\
 & \textbf{Qwen3-8B} & 42.5 & 10.0 & 17.2 & 23.2 \\
 & \textbf{Qwen3-4B} & 43.8 & 13.3 & 16.1 & 24.4 \\
 & \textbf{Qwen3-1.7B} & 42.5 & 6.7 & 9.5 & 19.6 \\
 & \textbf{Qwen3-0.6B} & 37.5 & 0.0 & 3.6 & 13.7 \\
\bottomrule
\end{tabular}
}
\end{table}

\begin{table}[tbp]
\caption{Language families and language codes supported by Qwen3 in Belebele Benchmark}
\label{tab:Belebele_language_codes}
\setlength\tabcolsep{2.5pt}
\adjustbox{center=\textwidth}{
\small
%
 \\
Sino-Tibetan    & 3        & zho\_Hans, mya\_Mymr, zho\_Hant                                                                                                                                                                                                                                                                       \\
Afro-Asiatic    & 8        & heb\_Hebr, apc\_Arab, acm\_Arab, ary\_Arab, ars\_Arab, arb\_Arab, mlt\_Latn, erz\_Arab                                                                                                                                                                                                                           \\
Austronesian    & 7        & ilo\_Latn, ceb\_Latn, tgl\_Latn, sun\_Latn, jav\_Latn, war\_Latn, ind\_Latn                                                                                                                                                                                                                                      \\
Dravidian       & 4        & mal\_Mlym, kan\_Knda, tel\_Telu, tam\_Taml                                                                                                                                                                                                                                                                       \\
Turkic          & 4        & kaz\_Cyrl, azj\_Latn, tur\_Latn, uzn\_Latn                                                                                                                                                                                                                                                                        \\
Tai-Kadai       & 2        & tha\_Thai, lao\_Laoo                                                                                                                                                                                                                                                                                               \\
Uralic          & 3        & fin\_Latn, hun\_Latn, est\_Latn                                                                                                                                                                                                                                                                                  \\
Austroasiatic   & 2        & vie\_Latn, khm\_Khmr                                                                                                                                                                                                                                                                                                \\
Other           & 7        & eus\_Latn, kor\_Hang, hat\_Latn, swh\_Latn, kea\_Latn, jpn\_Jpan, kat\_Geor  \\
\bottomrule
\end{tabular}
}
\end{table}

\begin{table}[tbp]
\caption{\textbf{Comparison of Belebele Benchmark performance between Qwen3 and other baseline models.} Scores are highlighted with the highest in \textbf{bold} and the second-best \underline{underlined}.}
\vspace{-1mm}
\label{tab:belebele}
\adjustbox{center=\textwidth}{
\small
\setlength{\tabcolsep}{2.5pt}
\scalebox{0.9}{
\begin{tabular}{@{}lcccccccccc@{}}
\toprule
Model                             & \tabincell{c}{\bf Indo-\\\bf European} & \tabincell{c}{\bf Sino- \\ \bf Tibetan} & \tabincell{c}{\bf Afro- \\ \bf Asiatic} & \textbf{Austronesian} & \textbf{Dravidian} & \textbf{Turkic} & \tabincell{c}{\bf Tai- \\ \bf Kadai} & \textbf{Uralic} & \textbf{Austroasiatic} & \textbf{Other}  \\
\midrule
Gemma-3-27B-IT                     & {\ul 89.2}           & 86.3                & \textbf{85.9}       & {\ul 84.1}          & 83.5             & {\ul 86.8}    & 81.0             & {\ul 91.0}    & 86.5                 & \textbf{87.0} \\
Qwen2.5-32B-Instruct               & 85.5                 & 82.3                & 80.4                & 70.6                & 67.8             & 80.8          & 74.5             & 87.0          & 79.0                 & 72.6          \\
QwQ-32B                            & 86.1                 & 83.7                & 81.9                & 71.3                & 69.3             & 80.3          & 77.0             & 88.0          & 83.0                 & 74.0          \\
\textbf{Qwen3-32B (Thinking)}      & \textbf{90.7}        & \textbf{89.7}       & {\ul 84.8}          & \textbf{86.7}       & \textbf{84.5}    & \textbf{89.3} & {\ul 83.5}       & \textbf{91.3} & \textbf{88.0}        & {\ul 83.1}    \\
\textbf{Qwen3-32B (Non-thinking)}  & 89.1                 & {\ul 88.0}          & 82.3                & 83.7                & {\ul 84.0}       & 85.0          & \textbf{85.0}    & 88.7          & \textbf{88.0}        & 81.3          \\ \midrule
Gemma-3-12B-IT                     & 85.8                 & {\ul 83.3}          & \textbf{83.4}       & 79.3                & {\ul 79.0}       & {\ul 82.8}    & 77.5             & {\ul 89.0}    & \textbf{83.0}        & {\ul 81.6}    \\
Qwen2.5-14B-Instruct               & 82.7                 & 78.9                & 80.4                & 69.1                & 66.2             & 74.2          & 72.2             & 83.9          & 77.9                 & 70.4          \\
\textbf{Qwen3-14B (Thinking)}      & \textbf{88.6}        & \textbf{87.3}       & {\ul 82.4}          & \textbf{82.4}       & \textbf{81.0}    & \textbf{83.8} & \textbf{83.5}    & \textbf{91.0} & {\ul 82.5}           & \textbf{81.7} \\
\textbf{Qwen3-14B (Non-thinking)}  & {\ul 87.4}           & 82.7                & 80.1                & {\ul 80.7}          & 78.0             & 81.8          & {\ul 80.5}       & 87.7          & 81.5                 & 77.0          \\ \midrule
Gemma-3-4B-IT                      & 71.8                 & 72.0                & 63.5                & 61.7                & 64.8             & {\ul 64.0}    & {\ul 61.5}       & 70.7          & 71.0                 & {\ul 62.6}    \\
Qwen2.5-3B-Instruct                & 58.0                 & 62.3                & 57.2                & 47.9                & 36.9             & 45.1          & 49.8             & 50.6          & 56.8                 & 48.4          \\
\textbf{Qwen3-4B (Thinking)}       & \textbf{82.2}        & \textbf{77.7}       & \textbf{74.1}       & \textbf{73.0}       & \textbf{74.3}    & \textbf{76.3} & \textbf{68.5}    & \textbf{83.0} & \textbf{74.5}        & \textbf{67.9} \\
\textbf{Qwen3-4B (Non-thinking)}   & {\ul 76.0}           & {\ul 77.0}          & {\ul 65.6}          & {\ul 65.6}          & {\ul 65.5}       & {\ul 64.0}    & 60.5             & {\ul 74.0}    & {\ul 74.0}           & 61.0          \\\midrule
Gemma-3-1B-IT                      & 36.5                 & 36.0                & 30.0                & 29.1                & 28.8             & 27.3          & 28.0             & 32.7          & 33.0                 & 30.9          \\
Qwen2.5-1.5B-Instruct              & 41.5                 & 43.0                & 39.6                & 34.8                & 28.6             & 29.7          & 39.4             & 33.8          & 42.0                 & 36.0          \\
\textbf{Qwen3-1.7B (Thinking)}     & \textbf{69.7}        & \textbf{66.0}       & \textbf{59.4}       & \textbf{58.6}       & \textbf{52.8}    & \textbf{57.8} & \textbf{53.5}    & \textbf{70.3} & \textbf{63.5}        & \textbf{53.4} \\
\textbf{Qwen3-1.7B (Non-thinking)} & {\ul 58.8}           & {\ul 62.7}          & {\ul 50.8}          & {\ul 53.0}          & {\ul 43.3}       & {\ul 48.0}    & {\ul 46.0}       & {\ul 54.3}    & {\ul 54.0}           & {\ul 43.9}    \\ 
\bottomrule
\end{tabular}
}
}
\end{table}

\clearpage
\bibliography{biblio}
\bibliographystyle{colm2024_conference}

\end{document}